\pdfoutput=1

\documentclass[11pt]{article}

\usepackage[final]{acl}

\usepackage{times}
\usepackage{latexsym}

\usepackage[T1]{fontenc}

\usepackage[utf8]{inputenc}

\usepackage{microtype}

\usepackage{inconsolata}

\usepackage{graphicx}

\usepackage{booktabs}
\usepackage{tabularx}
\usepackage{multicol}
\usepackage{multirow}
\usepackage{graphicx}
\usepackage{subfig}
\usepackage{float}
\usepackage{makecell}

\usepackage{longtable}

\usepackage{caption}
\usepackage{subcaption}
\usepackage{ltxtable}

\usepackage{xcolor}
\usepackage{xtab}
\renewcommand\fbox{\fcolorbox{gray}{white}}

\title{Are Large Language Models More Empathetic than Humans?}

\author{Anuradha Welivita, and Pearl Pu\\
  School of Computer and Communication Sciences \\
  École Polytechnique Fédérale de Lausanne \\
  Switzerland \\
  \texttt{\{kalpani.welivita,pearl.pu\}@epfl.ch}\\}

\begin{document}
\maketitle
\begin{abstract}


With the emergence of large language models (LLMs), investigating if they can surpass humans in areas such as emotion recognition and empathetic responding has become a focal point of research. This paper presents a comprehensive study exploring the empathetic responding capabilities of four state-of-the-art LLMs: \textit{GPT-4}, \textit{LLaMA-2-70B-Chat}, \textit{Gemini-1.0-Pro}, and \textit{Mixtral-8x7B-Instruct} in comparison to a human baseline. We engaged 1,000 participants in a between-subjects user study, assessing the empathetic quality of responses generated by humans and the four LLMs to 2,000 emotional dialogue prompts meticulously selected to cover a broad spectrum of 32 distinct positive and negative emotions. Our findings reveal a statistically significant superiority of the empathetic responding capability of LLMs over humans. GPT-4 emerged as the most empathetic, marking $\approx$31\% increase in responses rated as \textit{Good} compared to the human benchmark. It was followed by LLaMA-2, Mixtral-8x7B, and Gemini-Pro, which showed increases of approximately 24\%, 21\%, and 10\% in \textit{Good} ratings, respectively. We further analyzed the response ratings at a finer granularity and discovered that some LLMs are significantly better at responding to specific emotions compared to others. The suggested evaluation framework offers a scalable and adaptable approach for assessing the empathy of new LLMs, avoiding the need to replicate this study's findings in future research.

\end{abstract}

\section{Introduction}
\label{sec:introduction}

\begin{figure*}
  \centering
  \includegraphics[width=0.95\textwidth]{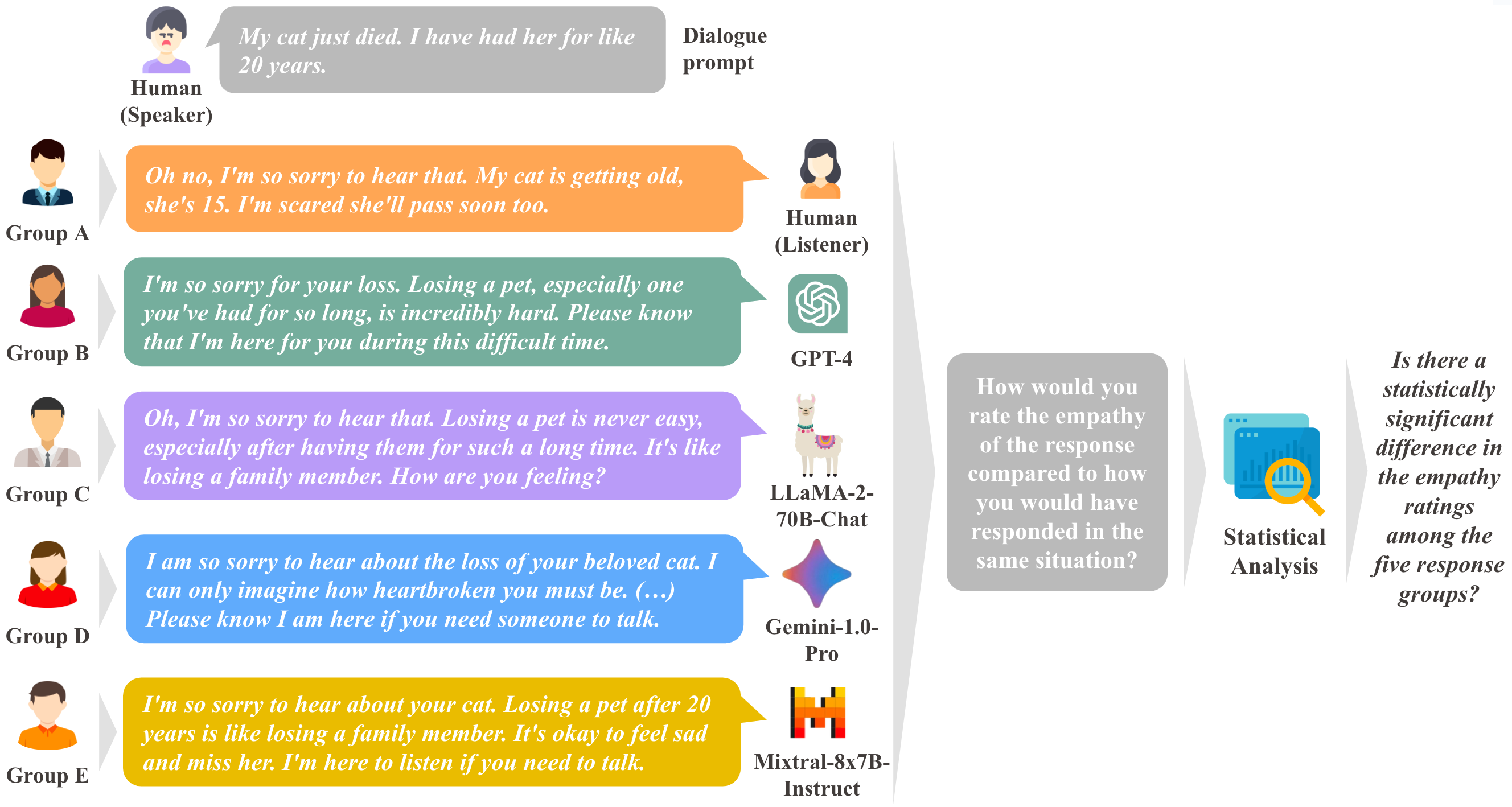}
  \caption{Between-subjects experiment design to evaluate the level of empathy demonstrated by LLMs compared to a human baseline when responding to emotional situations.}
  \label{fig:experiment_design}
\vspace{-2mm}
\end{figure*}


This era is marked by massive developments in artificial intelligence (AI), especially large language models (LLMs). They have exhibited performance exceeding humans across a variety of traditional language processing tasks such as question answering, text summarization, and commonsense reasoning \cite{laskarsystematic, ziyulens}. While there are public benchmarks and evaluation frameworks to evaluate LLMs' performance on these tasks, there is a lack of such resources to evaluate LLMs' ability to generate empathetic responses. Empathetic response generation requires generating replies that are not only contextually relevant and coherent but also demonstrate understanding, compassion, and emotional support towards the user's situation and feelings \cite{empatheticdialogues}. This is particularly challenging as empathy, being a deeply nuanced human experience, requires not only linguistic proficiency but also a deep understanding of human psychology, emotions, and social context \cite{ioannidou2008empathy}. 

Empathy is a multifaceted construct, encompassing cognitive, affective, and compassionate counterparts \cite{ekman2004emotions, decety2006human, powell2017situational}. Each component plays a crucial role in holistic empathetic engagement. Cognitive empathy is understanding and accurately identifying others’ feelings. Affective empathy is sharing the other person's emotions. Compassionate empathy is taking action to help the other person deal with their emotions. Empathy is a key component in making artificial conversational agents human-like, which fosters trust and rapport with the user \cite{liu2022artificial} and helps to increase people’s adoption of this technology \cite{goetz2003matching, stroessner2019social, svikhnushina2022peace}. So, evaluating the empathetic capabilities of LLMs that power artificial conversational agents plays a big role in deciding people’s willingness to use this technology.  




Existing studies that evaluate the empathetic capabilities of LLMs encompass major limitations. Most of them use automatic evaluation metrics that do not necessarily correlate with human perceptions of empathy \cite{belkhir2023beyond, loh2023harnessing}. Most evaluations are focused on the healthcare domain involving a lot of negative emotions \cite{chen2023llm, ayers2023comparing, liu2023leveraging}. But empathy plays an important role in responding to both positive and negative emotions encountered in daily conversations. Also, most studies investigate LLMs' ability to respond in general to emotions (which are mostly coarse-grained) as a whole, without analyzing them at a finer level \cite{lee2024large, zhao2023chatgpt, qian2023harnessing, lee2022does, fu2023reasoning, loh2023harnessing}. This makes it impossible to observe any variability in LLMs' performance when responding to diverse emotions. Last, but most importantly, all studies we came across used \textbf{within-subjects} study designs where the same participant evaluated responses generated by different models \cite{lee2024large, lee2022does, ayers2023comparing, fu2023reasoning, zhao2023chatgpt, qian2023harnessing}. In addition to introducing evaluation biases caused due to over-exposure to different model outputs and the order they are shown to the participants, this type of study design makes the evaluation approach not scalable to incorporate new and updated LLMs. 




Addressing the above limitations, we designed a \textbf{between-subjects} user study, recruiting 1,000 people from the crowdsourcing platform Prolific (\url{www.prolific.com}), in which 200 participants each were assigned to rate responses generated by humans and four state-of-the-art LLMs: \textit{GPT-4} \cite{OpenAI_2023}, \textit{LLaMA-2-70B-Chat} \cite{touvron2023llama}, \textit{Gemini-1.0-Pro} \cite{Pichai_2023}, and \textit{Mixtral-8x7B-Instruct} \cite{AI_2024} (see Figure \ref{fig:experiment_design}). We use 2,000 emotional dialogue prompts from the state-of-the-art EmpatheticDialogues dataset \cite{empatheticdialogues}, which contains chit-chat oriented human-human conversations, to form the human baseline required for our study as well as to initiate responses from the LLMs. We carefully selected the dialogue prompts to be equally distributed over a broad spectrum of 32 positive and negative emotions so that we can analyze whether there are any significant differences between humans and LLMs when responding to such distinct emotions. We prompt the four LLMs to generate a response to a given dialogue prompt, with instructions defining empathy in terms of its cognitive, affective, and compassionate counterparts. We adopt a simple and straightforward evaluation scale to gauge the empathy level in these responses. We perform rigorous statistical analysis to identify whether there are any statistically significant differences between the empathy ratings of humans and the four LLMs when responding to positive and negative emotional situations. We further delve into each finer emotion category and observe whether there are any significant differences in the way humans and LLMs respond to these individual emotions. Due to the careful and thorough design, this evaluation framework provides a robust and extensible solution to evaluate the empathetic quality of emerging LLMs without having to replicate the current study \footnote{The code for the evaluation framework and our dataset available at \url{github.com/anuradha1992/llms-vs-humans}}.

\section{Literature Review}



Different studies use different approaches to evaluate empathy in LLMs, most of which encompass automatic evaluation criteria. For example, Loh and Raamkumar \shortcite{loh2023harnessing} investigated the capability of five state-of-the-art LLMs including GPT-3.5, GPT-4, PaLM-2---the predecessor of Gemini, and Vicuna---based on LLaMA-1 to generate empathetic responses using $\approx$2,550 dialogue prompts from the EmpatheticDialogues dataset. They utilized three automatic empathy-related evaluation metrics: 1) Emotional Reactions (indicative of affective empathy); 2) Interpretations (indicative of cognitive empathy); and 3) Explorations (indicative of cognitive empathy) \cite{sharma-etal-2020-computational} and discovered that
LLMs' responses scored higher across the three metrics compared to those generated by traditional dialogue systems and humans. However, their evaluation is purely based on automatic evaluation, which does not necessarily correlate with how human users perceive the responses generated by the LLMs. A user-based evaluation could either validate the above observations or bring forth vastly different results. Belkhir and Sadat \shortcite{belkhir2023beyond} analyzed GPT-3.5's ability to produce empathetic responses, using precision, accuracy, and recall related to the emotion conveyed in the responses. However, empathetic communication does not always have to be emotional; it can sometimes be more neutral, focusing on specific intentions, as noted by Welivita and Pu \shortcite{welivitapu2020taxonomy}. This raises questions about the appropriateness of such metrics for evaluating empathetic responses.


Some studies have utilized questionnaires and psychological scales that are primarily designed to measure the empathy levels of humans on LLMs without considering their applicability. Schaaff et al. \shortcite{schaaff2023exploring} used standardized questionnaires from psychology such as Interpersonal Reactivity Index \cite{davis1980interpersonal}, Empathy Quotient \cite{lawrence2004measuring}, and Toronto Empathy Questionnaire \cite{spreng2009toronto} to assess the level of empathy exhibited by GPT-3.5 compared to humans. Elyoseph et al. \shortcite{elyoseph2023chatgpt} utilized the Levels of Emotional Awareness Scale (LEAS) \cite{lane1990levels} to evaluate GPT-3.5's ability to identify and describe emotions compared to the general population. But the applicability of this type of scales designed to evaluate humans' emotion understanding and empathy levels on LLMs is debatable. 


Research evaluating the empathetic responding ability of LLMs using human evaluators employ \textbf{within-subjects} designs, where the same participant evaluates different model outputs \cite{lee2024large, lee2022does, ayers2023comparing, fu2023reasoning, zhao2023chatgpt, qian2023harnessing}. For instance, Lee et al. \shortcite{lee2024large}, conducted a within-subjects study with 200 participants evaluating responses generated by humans, GPT-4, LLaMA-2, and Mixtral for 120 posts from Reddit. As elaborated in Section \ref{sec:introduction} this type of study is not extensible to newer and updated LLMs, requiring to reconduct the study from scratch when new LLMs emerge. Moreover, the relatively small sample size used fails to provide sufficient data to arrive at robust statistical conclusions. The above studies utilize standard A/B testing or a 5 or 7-point numerical rating scale (sometimes without any textual interpretations for each option) to rate the empathy-level of the responses generated by the LLMs. While effective in certain contexts, these methods have notable disadvantages. The rapid evolution of LLMs makes findings from A/B tests quickly outdated. The interpretation of scale points can vary widely among individuals, making it difficult to achieve consistent measurements across diverse participant groups. Most studies also lack a human baseline for comparison. This lack of a common ground to evaluate the empathetic responding capabilities of LLMs makes the evaluation complex and often not fully representative of how effective LLMs are in real-world interactions.

\section{The Dataset}

\begin{table*}[ht!]
\small
\centering
\begin{tabularx}{\textwidth}{X}
\toprule

\textit{Empathy is the ability to understand and share the feelings of another person. It is the ability to put yourself in someone else's shoes and see the world from their perspective.}\vspace{0.5mm}\\

\textit{Empathy is a complex skill that involves cognitive, emotional, and compassionate components.} \vspace{0.5mm}\\

\textit{\textbf{Cognitive empathy} is the ability to understand another person's thoughts, beliefs, and intentions. It is being able to see the world through their eyes and understand their point of view.} \vspace{0.5mm}\\

\textit{\textbf{Affective empathy} is the ability to experience the emotions of another person. It is feeling what they are feeling, both positive and negative.} \vspace{0.5mm}\\

\textit{\textbf{Compassionate empathy} is the ability to not only understand and share another person's feelings, but also to be moved to help if needed. It involves a deeper level of emotional engagement than cognitive empathy, prompting action to alleviate another's distress or suffering.} \vspace{0.5mm}\\

\textit{Empathy is important because it allows us to connect with others on a deeper level. It helps us to build trust, compassion, and intimacy. Empathy is also essential for effective communication and conflict resolution.} \vspace{0.5mm}\\

\textit{You are engaging in a conversation with a human. Respond in an empathetic manner to the following using on average 28 words and a maximum of 97 words.}\vspace{0.5mm}\\

\bottomrule
\end{tabularx}
\caption{The set of instructions used to prompt the large language models to generate empathetic responses.}
\label{tab:prompts}
\vspace{-2mm}
\end{table*}

To conduct the study, we used dialogues from the state-of-the-art EmpatheticDialogues dataset \cite{empatheticdialogues}, which consists of $\approx$25K dialogues spanning 32 fine-grained positive and negative emotions, selected from multiple annotation schemes, ranging from basic emotions derived from biological responses \cite{ekman,plutchik} to larger sets of subtle emotions derived from contextual situations \cite{skerry}. The dialogues in this dataset are curated by recruiting crowd workers from Amazon Mechanical Turk (AMT)\footnote{\url{https://www.mturk.com}}. The workers were paired together and were asked to role-play a dialogue, one person acting as the speaker and the other as the listener. The speaker was asked to pick an emotion, come up with a situation based on the chosen emotion, and start a conversation. The listener who is unaware of the emotion or the situation was asked to respond to the speaker. Based on the sample size predicted by power analysis (in Section \ref{sec:power_analysis}), we used randomly sampled 2,000 dialogues from this dataset, which are more or less equally distributed across the 32 emotions for our study (see Appendix \ref{app:emotions}). Though the dialogues spanned up to a maximum of 8 turns, for simplicity, we selected only the first two dialogue turns along with the emotion the dialogues were based on and the situation description. This formed the human baseline for our study. 

In one of our previous studies, we used two different prompts to instruct the LLM GPT-4 to generate responses given the 1\textsuperscript{st} turn of the dialogues. The first one was a very basic prompt that did not define the concept of empathy nor explicitly asked the model to generate an empathetic response. The second prompt defined the concept of empathy concerning its cognitive, affective, and compassionate counterparts and explicitly asked the model to respond in an empathetic manner to the given dialogue utterance. We observed that the one that utilized the second prompt outperformed the one that utilized the basic prompt as well as the human baseline with respect to the empathy ratings allocated by the study participants. Thus, we utilized the same empathy-defining instructions to prompt the LLMs compared in this study to generate responses. Table \ref{tab:prompts} denotes this set of instructions. 

For comparison with the human baseline, we use responses generated by four state-of-the-art LLMs: GPT-4 \cite{OpenAI_2023}; LLaMA-2-70B-Chat \cite{touvron2023llama}; Gemini-1.0-Pro \cite{Pichai_2023}; and Mixtral-8x7B-Instruct \cite{AI_2024}. Details regarding the four LLMs are in Appendix \ref{app:llms}. We first manually inspected a random set of responses generated by a large group of LLMs that included other LLMs such as PaLM-2 \cite{anil2023palm}, ChatGLM-3 \cite{zeng2022glm}, Vicuna-180B \cite{vicuna2023}, and Falcon-40B-Instruct \cite{falcon40b} and selected the LLMs that seemingly generated the highest quality responses to evaluate against the human baseline. Appendix \ref{app:stats} denotes the statistics of all the prompt-response pairs evaluated in the study.

\section{Experiment Design}


\subsection{Between-Subjects vs Within-Subjects}
\label{app:between_groups}


In our experiment design, which was structured as a \textbf{between-subjects study}, participants were divided into five groups. The first group assessed the empathetic quality of responses from humans to both positive and negative emotional scenarios. Each of the other four groups were assigned to evaluate empathy in responses generated by one of the four LLMs to the same emotional dialogue scenarios. This type of study design offers distinct advantages over a \textbf{within-subjects approach}. In within-subjects studies, as one person evaluates two or more model outputs, the evaluator's perception of empathy could be distorted by overexposure to model outputs resulting in a bias in their evaluations---commonly known as the \textit{carry-over effect}. For example, an averagely empathetic response may be judged more harshly by the evaluator if they have already seen an extremely empathetic response given by another model. This could also lead to \textit{order effects}, as the sequence in which model outputs are presented to the workers may influence how they assess the responses. \cite{shaughnessy2000research, charness2012experimental, montoya2023selecting}. Within-subjects studies also cannot accommodate seamless integration of outputs from newly developed language models. Such a study design necessitates starting from scratch every time a new model is introduced, making prior results obsolete. Conversely, a between-subjects study design, which employs different participants for assessing each model, offers the adaptability needed to evaluate emerging language models. This method facilitates an ongoing evaluation of language models' evolving empathy capabilities, making it a desirable option for such assessments.   



\subsection{Selection of the Rating Scale}
\label{sec:rating_scale}


When choosing a rating scale to evaluate the empathetic quality of responses, we faced a decision between a simpler 3-point scale with options \textit{Bad}, \textit{Okay}, \textit{Good} and a more detailed 5-point scale with options \textit{Bad}, \textit{Fair}, \textit{Okay}, \textit{Good}, and \textit{Excellent}. To determine the better option, we conducted a pilot study with 100 participants from Prolific. Each participant rated 10 responses using both scales. Half of the participants rated on a 3-point scale first and then on a 5-point scale (Group A) and the other half vice versa (Group B). We measured the agreement between raters in the above two groups using weighted Cohen's kappa \cite{cohen1968weighted}. The results showed that the 3-point scale achieved a kappa score of $0.2817$, indicating fair agreement, whereas the 5-point scale scored $0.1813$, indicating poor agreement. Additionally, we assessed how well the ratings from each scale correlated with scores from EPITOME \cite{epitome}, an automatic empathy evaluation tool. The 3-point scale ratings showed a low but better correlation of $0.1731$ with EPITOME's emotional reaction scores, compared to a negligible correlation of $0.0811$ for the 5-point scale. These findings indicate that although individual preferences for different scale types may vary subjectively, the 3-point scale more successfully maintains the accuracy of objective empathy measurements, resulting in evaluations that are both more reliable and consistent compared to those using the 5-point scale. This makes the 3-point scale a preferable choice for assessing empathy in responses, enhancing consistency among human raters, and alignment with automated tools.

\subsection{Task Design}
\label{app:task_design}

The five groups of participants for the study were recruited through the Prolific crowdsourcing platform (\url{www.prolific.com}). Past research has indicated that Prolific outperforms other crowdsourcing platforms like AMT, CloudResearch (\url{www.cloudresearch.com}), Dynata (\url{www.dynata.com}), and Qualtrics (\url{www.qualtrics.com}) in aspects such as worker attentiveness, integrity, understanding, and dependability \cite{peer2022data, douglas2023data}. Participants in the five groups were balanced across demographic criteria: gender (male and female); and age group (young adulthood [19 - 25 years]; middle adulthood [26 - 45 years]; late adulthood [46 - 64 years]; and older adulthood [65 years and above]). Additionally, a survey based on the Toronto Empathy Questionnaire (TEQ) \cite{spreng2009toronto} measured the empathy propensity of each participant, i.e., their natural predisposition to empathize with others. Subsequent analysis indicated that the inclination towards empathy was comparably distributed among the five groups, suggesting that participant conditions were uniformly matched across the board (see Appendix \ref{app:propensity}). Each participant evaluated randomly chosen 10 dialogue responses generated by the same model. The source of the responses, whether from a human or an LLM, was unknown to the participants. They were tasked with rating the empathy of the responses as either \textit{Bad}, \textit{Okay}, or \textit{Good}, relative to how they would have responded in similar situations. Furthermore, participants were introduced to the concept of empathy through a tutorial that covered its cognitive, affective, and compassionate dimensions. This tutorial was identical to the one used to prompt the LLMs and included exemplary dialogues from the EmpatheticDialogues dataset. These examples were chosen based on high ratings of empathy, relevance, and fluency by the human participants involved in the dataset's creation.

\subsection{Quality Control}
\label{app:quality_control}




To ensure a high standard of data quality, our study selectively recruited participants who were proficient in English and had a track record of at least 100 prior submissions with an approval rate exceeding 95\%. Following the selection criteria, the Toronto Empathy Questionnaire (TEQ), which was used to measure the workers' propensity to empathize, contained 8 reserve scale questions. These questions were used to gauge the quality of the workers and their attentiveness to the task. 

\subsection{Statistical Test and Sample Size}
\label{sec:power_analysis}

To analyze the results from the study we use the \textbf{chi-square test of independence} \cite{mchugh2013chi} that tests whether there is any statistically significant difference between the proportion of \textit{Bad}, \textit{Okay}, and \textit{Good} ratings of the five response groups. When analyzing categorical ratings, particularly if the data involves ratings from different groups (like humans vs LLMs), the chi-Square test of independence is often a strong choice due to its robustness and the straightforward interpretability of the results \cite{field2013discovering}. The null and the alternate hypotheses of this statistical test are as below.


\begin{table}[H]
\small
\centering
\begin{tabularx}{\linewidth}{X}

\textbf{${\chi}^2$ test of independence:}\vspace{-1.5mm}
\begin{itemize}
\setlength\itemsep{0.25em}
\item \textbf{Null hypothesis:} There is no difference between the proportion of \textit{Bad}, \textit{Okay}, and \textit{Good} ratings of the five groups of responses.
\item \textbf{Alternative hypothesis:} There is a difference between the proportion of \textit{Bad}, \textit{Okay}, and \textit{Good} ratings of at least one out of the five groups of responses.\vspace{-2mm}

\end{itemize}

\end{tabularx}
\label{table:hypothesis}
\vspace{-3mm}
\end{table}

We used the G-Power software \cite{faul2009statistical} to compute the minimal sample size required to detect a significant difference between the ratings of the five response groups. For the chi-square test of independence with a medium effect size $(0.3)$, a significance level $(\alpha)$ of $0.05$, and a power $(1-\beta)$ of $0.95$, the minimal total sample size required is $253$ (i.e. at least $51$ participants per group). When statistically analyzing the differences in empathy ratings when responding to positive and negative emotions separately, the minimal sample size required becomes twice the sample size suggested above (i.e. at least $102$ participants per group). From a prior study, we had already engaged $200$ participants to evaluate responses generated by humans and GPT-4. To ensure compatibility, we additionally recruited $600$ participants from Prolific to rate responses generated by the LLMs: LLaMA-2; Gemini-Pro; and Mixtral-8x7B. That is $200$ participants per group, which is sufficiently above the minimal sample size. One participant was asked to rate $10$ responses. Altogether our study compares empathy ratings received for $10,000$ responses ($2,000$ responses per group).  



\section{Results}

\begin{figure*}
  \centering
  \includegraphics[width=0.8\textwidth]{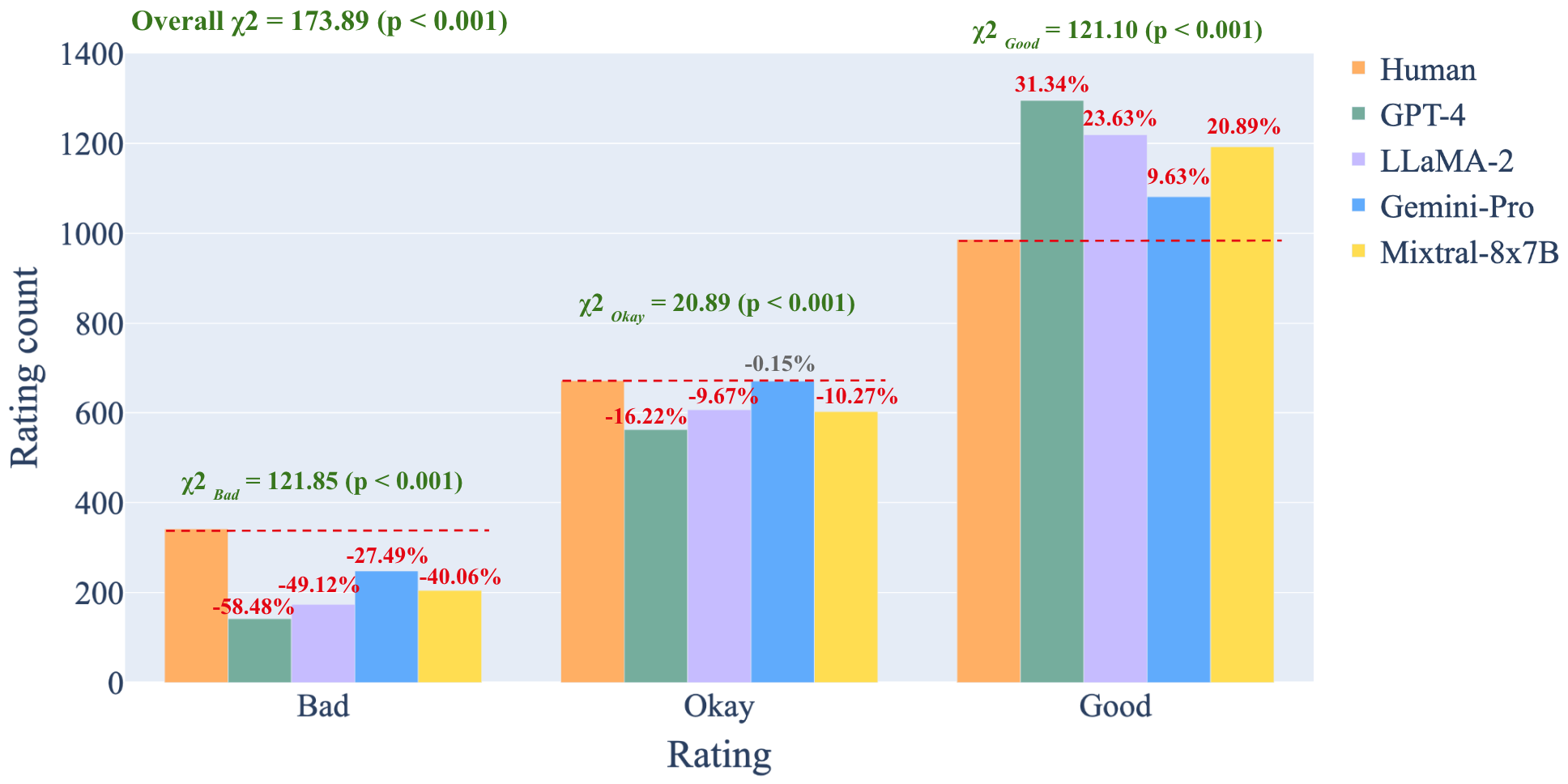}

  \caption{The \textit{Good}, \textit{Okay}, and \textit{Bad} rating counts corresponding to the responses generated by humans, GPT-4, LLaMA-2, Gemini-Pro, and Mixtral-8x7B. The percentage gains of the LLMs' response ratings compared to the humans' response ratings are indicated at the top of each bar. The gains indicated in red are statistically significant.}

  
  \label{fig:main_results}
  \vspace{-2mm}
\end{figure*}

Figure \ref{fig:main_results} visualizes the number of \textit{Good}, \textit{Okay}, and \textit{Bad} ratings received by the responses generated by humans, and the four LLMs for dialogue prompts spanning across all emotions as a whole. The ${\chi}^2$ and the p-values obtained by applying the chi-square test of independence to the number of \textit{Good}, \textit{Okay}, and \textit{Bad} ratings collectively and for each category independently indicated that there is a statistically significant difference between the proportion of \textit{Good}, \textit{Okay}, and \textit{Bad} ratings of at least one out of the five response groups. We computed the percentage gains of the ratings received by each LLM compared to the human baseline under each rating category. GPT-4 was observed to generate the most empathetic responses with $\approx31\% ({\chi}^2 = 96.77, p < .001)$ gain in the number of \textit{Good} ratings compared to the humans. LLaMA-2, Mixtral-8x7B, and Gemini-Pro were observed to follow GPT-4 with $\approx24\% ({\chi}^2 = 54.40, p < .001)$, $\approx21\% ({\chi}^2 = 42.36, p < .001)$, and $\approx10\% ({\chi}^2 = 8.85, p < .01)$ gain in the number of \textit{Good} ratings, respectively, compared to the human baseline. Note that when calculating the ${\chi}^2$ values here, we considered \textit{Good} ratings as one category and combined \textit{Bad} and \textit{Okay} ratings as another category.   



\begin{figure*}
     
     \centering
     \subfloat[\centering Positive dialogues]{{\includegraphics[width=0.455\textwidth]{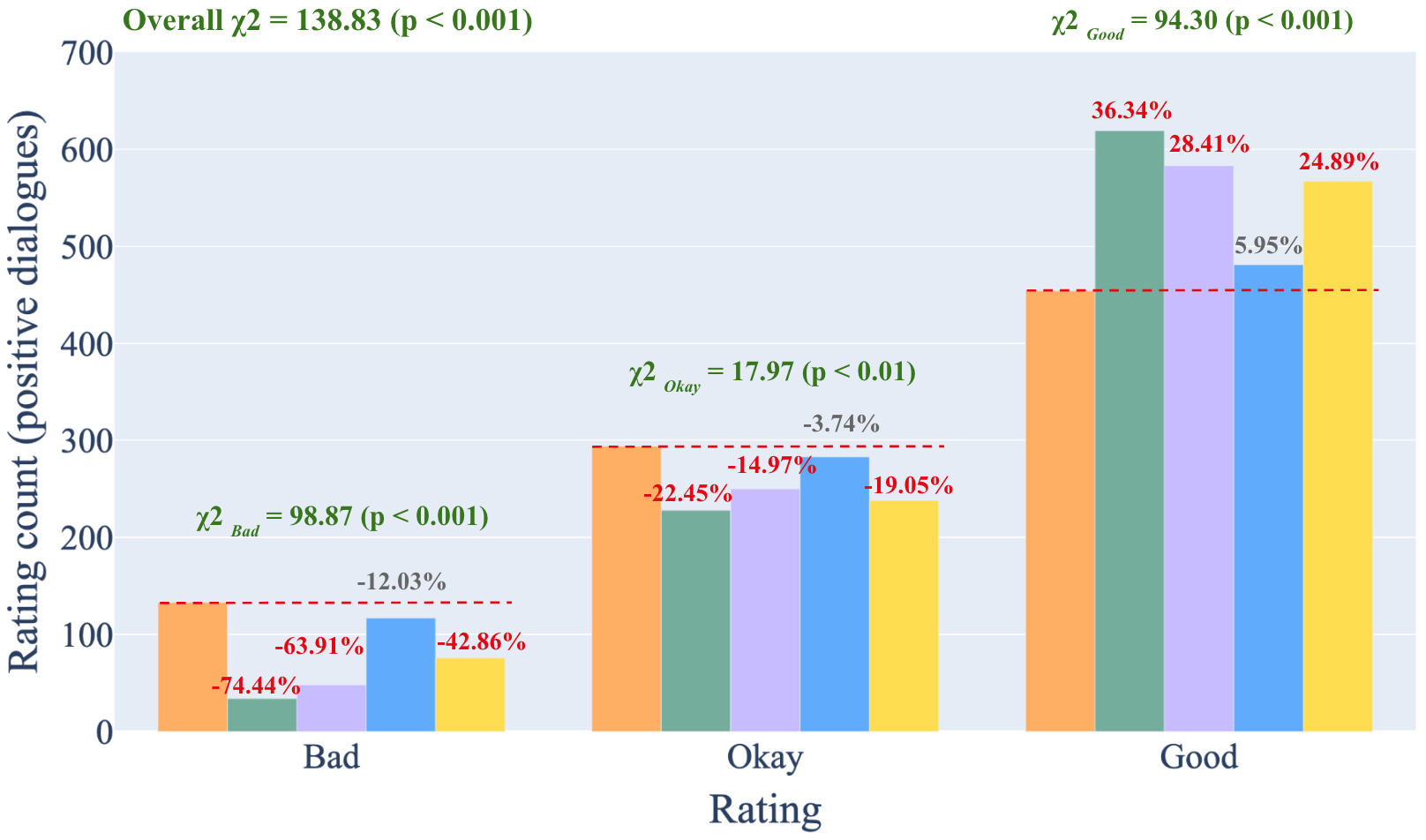}}}     
     \subfloat[\centering Negative dialogues]{{\includegraphics[width=0.54\textwidth]{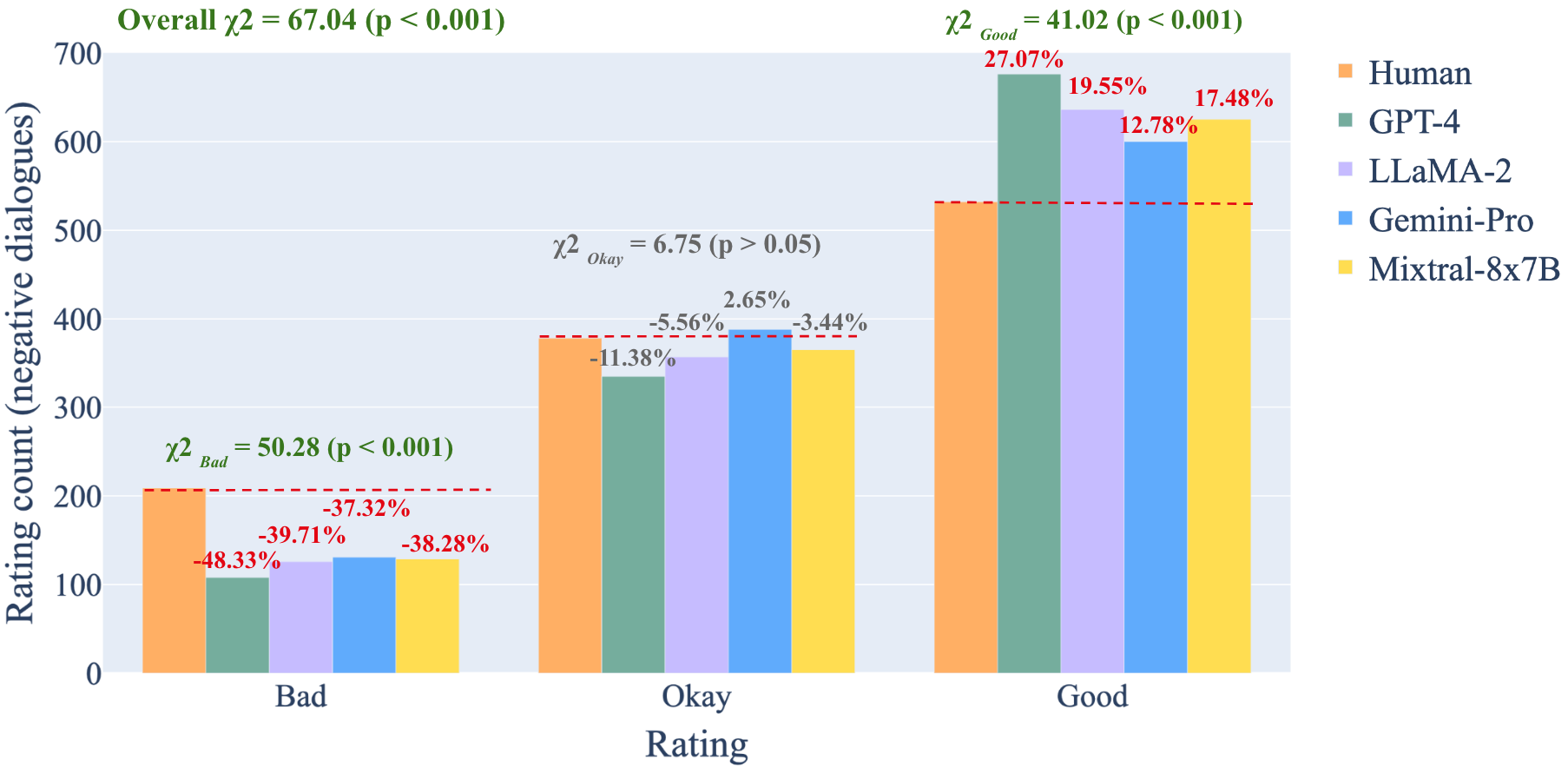}}}
     
     \caption{The \textit{Good}, \textit{Okay}, and \textit{Bad} rating counts corresponding to the responses generated by humans, GPT-4, LLaMA-2, Gemini-Pro, and Mixtral-8x7B for positive and negative emotional dialogue prompts.}
     \label{fig:pos_neg_results}
\vspace{-2mm}
\end{figure*}

Figure \ref{fig:pos_neg_results} visualizes the number of \textit{Good}, \textit{Okay}, and \textit{Bad} ratings received by the responses generated by humans and the four LLMs for positive and negative emotional dialogue prompts, separately. All four LLMs outperformed the human baseline across both positive and negative emotions in the number of \textit{Good} ratings received. Here too, GPT-4 ranked the top in the number of \textit{Good} ratings, obtaining percentage gains of $\approx36\% ({\chi}^2 = 64.10, p < .001)$ and $\approx27\% ({\chi}^2 = 36.78, p < .001)$, respectively across positive and negative emotions, compared to the human baseline. LLaMA-2 and Mixtral-8x7B followed GPT-4 when responding to positive emotions obtaining $\approx28\% ({\chi}^2 = 38.40, p < .001)$, and $\approx25\% ({\chi}^2 = 29.21, p < .001)$ gain in the number of \textit{Good} ratings, respectively, compared to the human baseline. However, the percentage gain in the number of \textit{Good} ratings obtained by Gemini-Pro across positive emotions was observed to be statistically insignificant compared to those received by the human responses $(\uparrow = 5.95\%, {\chi}^2 = 1.54, p > .05)$. LLaMA-2, Mixtral-8x7B, and Gemini-Pro followed GPT-4 when responding to negative emotions obtaining $\approx20\% ({\chi}^2 = 19.0, p < .001)$, $\approx17\% ({\chi}^2 = 15.15, p < .001)$, and $\approx13\% ({\chi}^2 = 8.02, p < .01)$ gain in the number of \textit{Good} ratings, respectively, compared to the human baseline. 




Further, we computed the percentage gains of the categorical ratings received by each LLM compared to the human baseline when responding to each of the 32 positive and negative emotions (See Table \ref{tab:neg_break} in Appendix \ref{app:emotion_breakdown}). This finer analysis allowed us to observe that some LLMs are significantly better than humans when responding to specific emotions compared to others. It could be observed that GPT-4 obtains statistically significant percentage gains in the number of \textit{Good} ratings over the human baseline across most positive emotion categories such as
\textit{Impressed} $(\uparrow = 56\%, {\chi}^2 = 10.62, p < .01)$, \textit{Surprised} $(\uparrow = 79\%, {\chi}^2 = 10.33, p < .01)$, \textit{Grateful} $(\uparrow = 65\%, {\chi}^2 = 8.36, p < .01)$, \textit{Proud} $(\uparrow = 50\%, {\chi}^2 = 7.7, p < .01)$, \textit{Confident} $(\uparrow = 44\%, {\chi}^2 = 6.86, p < .01)$, \textit{Joyful} $(\uparrow = 42\%, {\chi}^2 = 6.34, p < .05)$, and \textit{Excited} $(\uparrow = 47\%, {\chi}^2 = 5.41, p < .05)$; LLaMA-2 across emotions \textit{Grateful} $(\uparrow = 65\%, {\chi}^2 = 8.36, p < .01)$, \textit{Surprised} $(\uparrow = 71\%, {\chi}^2 = 8.14, p < .01)$, \textit{Proud} $(\uparrow = 44\%, {\chi}^2 = 5.69, p < .05)$, \textit{Excited} $(\uparrow = 44\%, {\chi}^2 = 4.59, p < .05)$, \textit{Hopeful} $(\uparrow = 39\%, {\chi}^2 = 4.27, p < .05)$, and \textit{Prepared} $(\uparrow = 39\%, {\chi}^2 = 4.06, p < .05)$; and Mixtral-8x7B across emotions \textit{Proud} $(\uparrow = 59\%, {\chi}^2 = 11.44, p < .001)$, \textit{Grateful} $(\uparrow = 58\%, {\chi}^2 = 6.36, p < .05)$, and \textit{Excited} $(\uparrow = 47\%, {\chi}^2 = 5.41, p < .05)$. 

Compared to the positive emotions, we could only observe the four LLMs obtaining significant gains in the number of \textit{Good} ratings over humans only across a few negative emotions such as \textit{Afraid} (LLaMA: $\uparrow = 50\%, {\chi}^2 = 4.66, p < .05$; GPT: $\uparrow = 46\%, {\chi}^2 = 3.91, p < .05$), \textit{Apprehensive} (GPT: $\uparrow = 104\%, {\chi}^2 = 20.72, p < .001$; Gemini: $\uparrow = 60\%, {\chi}^2 = 6.23, p < .05$; LLaMA: $\uparrow = 52\%, {\chi}^2 = 4.57, p < .05$), \textit{Anxious} (GPT: $\uparrow = 75\%, {\chi}^2 = 9.2, p < .01$; LLaMA: $\uparrow = 63\%, {\chi}^2 = 6.22, p < .05$; Gemini: $\uparrow = 63\%, {\chi}^2 = 6.22, p < .05$; Mixtral: $\uparrow = 50\%, {\chi}^2 = 3.85, p < .05$), and \textit{Annoyed} (GPT: $\uparrow = 59\%, {\chi}^2 = 6.62, p < .05)$; Mixtral: $\uparrow = 52\%, {\chi}^2 = 4.97, p < .05)$). This implies that there is more room for these LLMs to improve their performance across other important negative emotion categories that commonly occur in day-to-day conversations.  






\section{Case Study}

Table \ref{tab:examples} shows an example, in which the response generated by the human was rated \textit{Bad} whereas the responses generated by the four LLMs were rated \textit{Good} by the participants. It could be noted that in the human response, the human responder speaks about themselves rather than focussing on the emotion of the speaker. On the other hand, all the four LLMs seem to recognize the emotion of the speaker and the love the speaker's grandmother has towards them and validate it using phrases such as \textit{That’s so sweet}, \textit{That’s so thoughtful}, \textit{That’s so heartwarming to hear!}, and \textit{Your grandmother’s thoughtfulness warms my heart}. What follows in the LLMs' responses are more complex reflections of what the speaker has said, which not only demonstrates understanding but also adds depth to the conversation, potentially leading to a more profound continuation of the dialogue. More such examples are denoted in Appendix \ref{app:examples}.  

\begin{table*}[ht!]
\small
\centering
\begin{tabularx}{\textwidth}{p{1.1cm} X}
\toprule

\textbf{Prompt} & \textit{My grandma sent a lot of food to me through my mom. She's still making sure I get good food to eat.} \textbf{(Sentimental)}\\


\multicolumn{2}{l}{\textbf{\underline{Responses:}}}\vspace{1mm}\\

\textbf{Human} & \textit{Some habits die hard.  My mom's always trying to send me food still. I'm a grown man.} \textbf{(Bad)}\\

\textbf{GPT} & \textit{That's so sweet of your grandma. It shows how much she cares about you. It must be comforting to know you're always on her mind.} \textbf{(Good)}\\

\textbf{LLaMA} & \textit{That's so thoughtful of your grandma to make sure you're well taken care of. It's great that she's still involved in your life and wants to ensure your well-being.} \textbf{(Good)}\\

\textbf{Gemini} & \textit{That's so heartwarming to hear! Your grandma's thoughtfulness shows how much she cares about you. It's wonderful that she's going to such efforts to ensure you're well-fed. I can imagine the love and warmth that comes with each meal. Thanks for sharing this special gesture.} \textbf{(Good)}\\

\textbf{Mixtral} & \textit{Your grandmother's thoughtfulness warms my heart. It's evident that she cares deeply for you and wants to ensure your well-being. Her actions speak volumes about the love and concern she holds for you.} \textbf{(Good)}\\

\bottomrule
\end{tabularx}
\caption{An example where the response generated by the human was rated \textit{Bad} whereas the responses generated by the LLMs were rated \textit{Good} by the participants.}
\label{tab:examples}
\vspace{-2mm}
\end{table*}

\section{Discussion}


The responses generated by all four LLMs surpassed the human responses in terms of empathetic quality by a statistically significant margin across all emotions as a whole, and across positive (except Gemini-pro) and negative emotions separately. Even though Gemini-Pro reported a significant gain ($\approx13\%$) compared to the human baseline across negative emotions, it failed to report statistically significant gains compared to the human baseline across positive emotions, which leaves more room for Gemini-Pro to improve its performance when responding to positive emotions. It was seen that LLMs GPT-4, LLaMA-2, and Mixtral-8x7B perform significantly better than humans when responding to most positive emotion categories such as \textit{Grateful}, \textit{Proud}, and \textit{Excited}. However, compared to positive emotions, the four LLMs could only perform significantly better than humans when responding to a few negative emotions such as \textit{Apprehensive} and \textit{Anxious}. This implies that the LLMs, while advanced in their empathetic understanding and response generation, have differential performance based on the valence of the emotions they are responding to. This could be due to a variety of factors such as the data the models are trained on, which may contain richer or more nuanced examples of responses to specific emotions, allowing the LLMs to learn more effective response strategies for these emotions.

The disparity in performance between positive and negative emotions also suggests that future iterations of these LLMs could benefit from targeted improvements in understanding and responding to more negative emotions. This could involve incorporating more diverse and nuanced examples of negative emotional responses into the training data or refining the models' algorithms to better capture the subtleties of negative emotional expressions.

Furthermore, the fact that LLMs outperform humans in empathetic response quality, especially in certain emotions, underscores the potential of these models in applications requiring emotional intelligence, such as mental health support, customer service, and social interactions. However, the variability in performance across different types of emotions also highlights the importance of ongoing research and development to enhance the models' emotional intelligence across the full spectrum of human emotions.

Overall, this study contributed knowledge on how empathy is conveyed in responses generated by different LLMs to diverse positive and negative emotional stimuli, compared to a human baseline. Due to the between-groups study design and the release of the dataset, the evaluation framework that we introduce could be extended to evaluate the empathetic responding capabilities of newer and updated versions of LLMs as they emerge.

\section{Limitations}

The choice of using a 3-point scale rather than a 5 or 7-point scale can raise questions regarding the granularity of the evaluation. As described in section \ref{sec:rating_scale}, given that the study involves a large and diverse pool of participants, we believe that the advantages of using a simple, straightforward, and a human-centric rating scale, outweigh the concerns regarding granularity. As is evident from our results, the 3-point scale, while offering less granularity than a 5 or 7-point scale, still provides sufficient variability to perform robust statistical analyses. This scale was adequate to reveal significant differences in human and LLM-generated empathetic responses, confirming its effectiveness in the context of our research objectives. This establishes a foundational benchmark for evaluating the empathetic quality of responses, serving as a stepping stone for more detailed future studies.

\section{Ethical Considerations}

\textbf{Data usage: } The study utilized dialogue prompt-response pairs from the state-of-the-art EmpatheticDialogues dataset \cite{empatheticdialogues}, which contains ethically sourced dialogues and is available publicly under the CC BY-NC 4.0 license. The dataset itself is anonymized to protect the privacy of individuals who contributed to its creation. We plan to publicly release the new artifacts generated in this study, including the responses from the four LLMs and the participants' empathy ratings, under the CC BY-NC 4.0 license. This licensing allows other researchers to modify, enhance, and further build upon our work for non-commercial purposes. By doing so, we aim to facilitate ongoing comparisons with newer and updated versions of LLMs, eliminating the need to replicate the entire study from the beginning.  



\textbf{Human experiment: } The human participants recruited from the crowdsourcing platform Prolific (\url{www.prolific.com}) were paid \texteuro2.25 for rating 10 responses that took on average 11 minutes and 23 seconds to complete. This was $\approx$1.3 times above the wage recommended as \textit{Good} (\texteuro9 per hour) by the Prolific crowdsourcing platform. All participants were informed about the purpose of the study and the nature of the tasks they would perform. The ratings were collected at the end of the task after the participants decided to submit their work. Intermediate annotations were not recorded. The participants were allowed to leave the task at any time without submitting their ratings. Random subsets of dialogue prompt-response pairs used in the study were manually inspected to ensure that the tasks assigned to the crowd workers were not psychologically distressing or offensive. In addition, efforts were made to recruit a diverse group of participants considering factors of gender and the age group that represent the broader population to avoid bias in the results.





\textbf{Transparency and reproducibility of the study: } The dialogue prompt-response pairs that were subjected to evaluation along with the participants' evaluations of these responses will be released publicly to ensure the transparency and reproducibility of our study. 

\textbf{Ethical concerns surrounding empathetic LLMs: } Given the black-box nature of LLMs and their limited controllability and interpretability, one should take caution when using them, particularly in sensitive application domains such as mental health and crisis support. The opaque nature of these models can lead to outputs that are unpredictable or misaligned with human expectations, which can raise significant ethical concerns. Also, LLM-generated responses can represent societal biases and discriminations that are inherently present in the training data, which can lead to discriminatory or unethical outputs. Thus, an ethical approach to deploying such LLMs in sensitive domains should involve rigorous checking for biases and continuously monitoring their performance across underrepresented social groups. Some research studies point out that over-reliance on AI for empathetic interactions could affect human empathy skills and alter traditional social interactions \cite{chen2024feels}. There is also a concern regarding the sincerity of the LLM-generated empathetic responses since LLMs cannot feel the users' emotions \cite{bove2019empathy}. Hence, it is important to be transparent about the nature of the LLM-generated responses to avoid over-reliance or emotional attachment to these agents over time.

\bibliography{acl_latex}

\appendix



\section{Distribution of Emotions}
\label{app:emotions}

\begin{figure*}
\centering
  \includegraphics[width=0.85\textwidth]{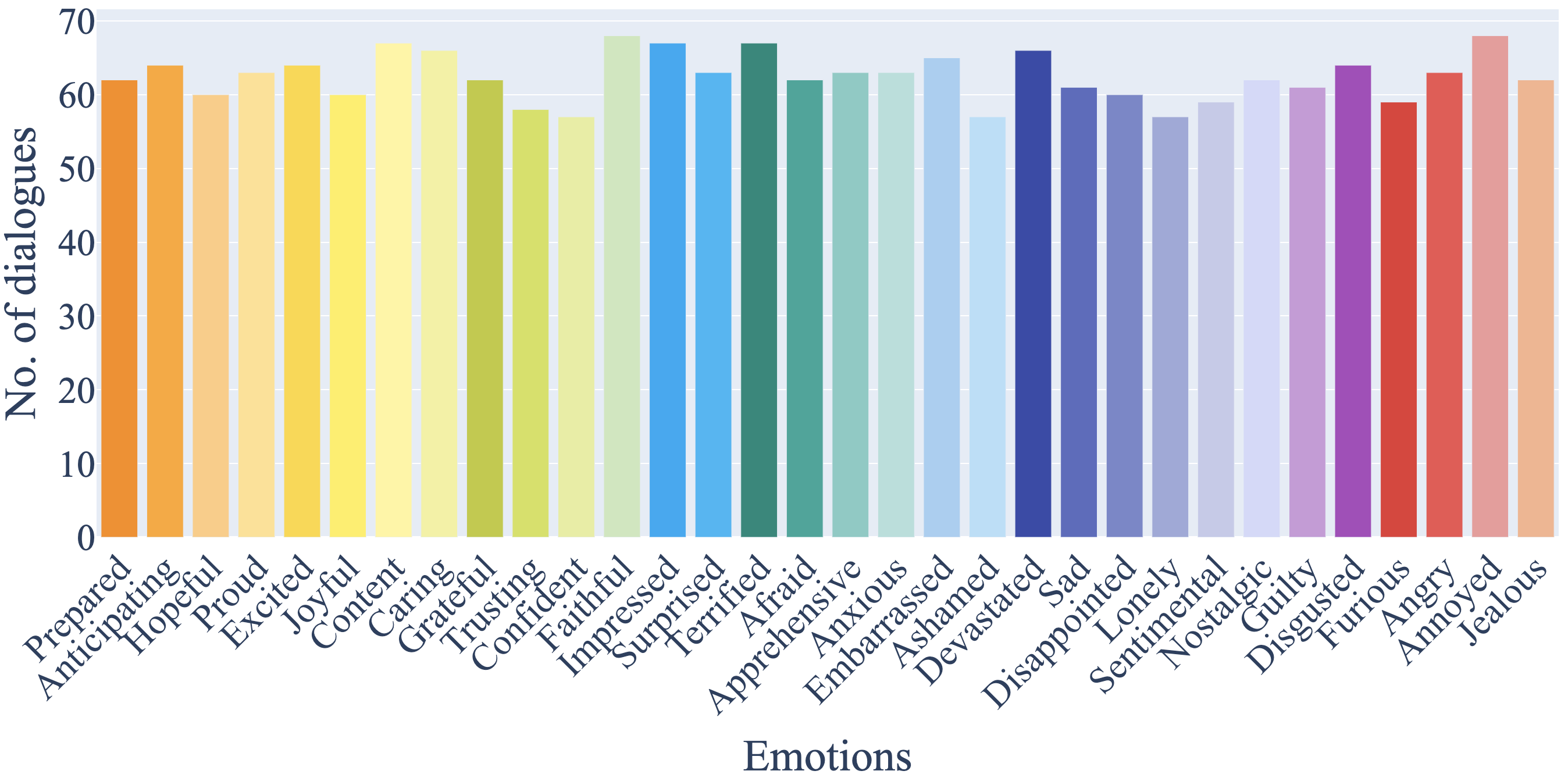}
 \caption{Distribution of the dialogue prompt-response pairs sampled from the EmpatheticDialogues dataset across the 32 positive and negative emotions.}
 \label{fig:emotions}
\end{figure*}

Figure \ref{fig:emotions} shows the distribution of the dialogue prompt-response pairs sampled from the EmpatheticDialogues dataset across the 32 positive and negative emotions. Table \ref{tab:emotions} shows the counts and the percentages of dialogue prompt-response pairs in the dataset corresponding to each emotion. It can be noted that the prompt-response pairs are more or less equally distributed across the 32 emotions. 

\begin{table}[ht!]
\small
\centering
\begin{tabular}{l c c}
\toprule

\textbf{Emotion} & \textbf{\# dialogues} & \textbf{\% of dialogues} \\

\midrule

\textbf{\underline{Positive emotions:}} & \textbf{881} & \textbf{44.05\%} \vspace{1mm} \\

Prepared & 62 & 3.10\% \\
Anticipating & 64 & 3.20\% \\
Hopeful & 60 & 3.00\% \\
Proud & 63 & 3.15\% \\
Excited & 64 & 3.20\% \\
Joyful & 60 & 3.00\% \\
Content & 67 & 3.35\% \\
Caring & 66 & 3.30\% \\
Grateful & 62 & 3.10\% \\
Trusting & 58 & 2.90\% \\
Confident & 57 & 2.85\% \\
Faithful & 68 & 3.40\% \\
Impressed & 67 & 3.35\% \\
Surprised & 63 & 3.15\% \vspace{1mm} \\

\textbf{\underline{Negative emotions:}} & \textbf{1119} & \textbf{55.95\%} \vspace{1mm} \\

Terrified & 67 & 3.35\% \\
Afraid & 62 & 3.10\% \\
Apprehensive & 63 & 3.15\% \\
Anxious & 63 & 3.15\% \\
Embarrassed & 65 & 3.25\% \\
Ashamed & 57 & 2.85\% \\
Devastated & 66 & 3.30\% \\
Sad & 61 & 3.05\% \\
Disappointed & 60 & 3.00\% \\
Lonely & 57 & 2.85\% \\
Sentimental & 59 & 2.95\% \\
Nostalgic & 62 & 3.10\% \\
Guilty & 61 & 3.05\% \\
Disgusted & 64 & 3.20\% \\
Furious & 59 & 2.95\% \\
Angry & 63 & 3.15\% \\
Annoyed & 68 & 3.40\% \\
Jealous & 62 & 3.10\% \\

\bottomrule
\end{tabular}
\caption{The counts and percentages of dialogue prompt-response pairs in the dataset corresponding to each emotion.}
\label{tab:emotions}
\end{table}

\section{Large Language Models}
\label{app:llms}

The study evaluated four state-of-the-art LLMs: GPT-4; LLaMA-2-Chat-70B; Gemini-1.0-Pro; and Mixtral-8x7B-Instruct. The details of the four LLMs are as follows.  

\vspace{2mm} 

\noindent\textbf{GPT-4} \cite{OpenAI_2023} developed by OpenAI (\url{openai.com}) is the latest model in their GPT series with an estimated 1.76 trillion parameters. GPT-4 is claimed to be more reliable, creative, and able to handle much more nuanced instructions than its predecessor GPT-3.5. The model considerably outperforms existing LLMs, alongside most state-of-the-art models which include benchmark-specific crafting or additional training protocols.

\vspace{2mm} 

\noindent\textbf{LLaMA-2-Chat-70B} \cite{touvron2023llama} developed by Meta AI (\url{ai.meta.com}), is an open-source LLM pre-trained on publicly available online data sources and fine-tuned on publicly available instruction tuning data \cite{chung2022scaling},  aligning the LLM towards dialogue-style instructions. We used the largest variant of LLaMA-2 with 70 billion parameters for this study.



\vspace{2mm}

\noindent\textbf{Gemini-1.0-Pro} \cite{Pichai_2023} developed by Google is a multimodal LLM trained to recognize and understand text, images, audio, and video. While Google does not reveal the exact number of parameters of this model and the data the model is trained on, it is developed based on the transformer architecture and relies on strategies like pre-training and fine-tuning, much as other LLMs. Independent research found that Gemini-1.0-Pro trails GPT-3.5-turbo across many of the traditional NLP benchmarks \cite{akter2023depth}. 

\vspace{2mm}



\noindent\textbf{Mixtral-8x7B-Instruct} \cite{AI_2024} developed by Mistral AI (\url{mistral.ai}), is a high-quality sparse mixture of experts model (SMoE) with 46.7B total parameters. The \textit{Instruct} model has been optimised through supervised fine-tuning and direct preference optimisation for careful instruction following. It is claimed to outperform LLaMA-2 (70B) on most traditional NLP benchmarks with 6x faster inference. The model is also claimed to match or outperform GPT-3.5 on most standard benchmarks.

\section{Statistics of the dialogue prompts and responses}
\label{app:stats}

Table \label{tab:stats} denotes the statistics of all the prompt-response pairs evaluated in the study.

\begin{table}[ht!]
\small
\centering
\begin{tabularx}{\linewidth}{X r r}
\toprule

\textbf{Model} & \textbf{Avg \# tokens} & \textbf{Max \# tokens} \\

\midrule

Dialogue prompt & 23.24 & 143\vspace{1mm}\\

\multicolumn{3}{l}{\textbf{\underline{Responses:}}\vspace{1mm}} \\

Human & 28.37 & 97 \\
GPT-4 & 34.94 & 65 \\
LLaMA-2-Chat-70B & 53.45 & 90 \\
Gemini-1.0-Pro & 53.99 & 93 \\
Mixtral-7x8B-Instruct & 61.35 & 95 \\

\bottomrule
\end{tabularx}
\caption{Statistics of the dialogue prompts and responses used for the study. The dialogue prompt here means the first dialogue utterance that initiates a reply. NLTK's tokenized package\protect\footnotemark was used to tokenize the text.}
\label{tab:stats}
\end{table}

\footnotetext{\url{https://www.nltk.org/api/nltk.tokenize.html}}

\section{Toronto Empathy Questionnaire}
\label{app:teq}

Table \ref{tab:teq} shows the questions in the Toronto Empathy Questionnaire (TEQ) \cite{spreng2009toronto} that were asked from the participants. Responses to the questions are scored according to the following scale for positively worded questions: Never = 0; Rarely = 1; Sometimes = 2; Often = 3; Always = 4. The negatively worded questions indicated are reverse-scored. Scores are summed to derive one's propensity to empathize.

\begin{table*}[ht!]
\small
\centering
\begin{tabular}{r l}
\toprule


1. & \textit{When someone else is feeling excited, I tend to get excited too} \\
2. & \textit{Other people’s misfortunes do not disturb me a great deal\textsuperscript{*}} \\
3. & \textit{It upsets me to see someone being treated disrespectfully} \\
4. & \textit{I remain unaffected when someone close to me is happy\textsuperscript{*}} \\
5. & \textit{I enjoy making other people feel better} \\
6. & \textit{I have tender, concerned feelings for people less fortunate than me} \\
7. & \textit{When a friend starts to talk about his or her problems, I try to steer the conversation towards something else\textsuperscript{*}} \\
8. & \textit{I can tell when others are sad even when they do not say anything} \\
9. & \textit{I find that I am “in tune” with other people’s moods} \\
10. & \textit{I do not feel sympathy for people who cause their own serious illnesses\textsuperscript{*}} \\
11. & \textit{I become irritated when someone cries\textsuperscript{*}} \\
12. & \textit{I am not really interested in how other people feel\textsuperscript{*}} \\
13. & \textit{I get a strong urge to help when I see someone who is upset} \\
14. & \textit{When I see someone being treated unfairly, I do not feel very much pity for them\textsuperscript{*}} \\
15. & \textit{I find it silly for people to cry out of happiness\textsuperscript{*}} \\
16. & \textit{When I see someone being taken advantage of, I feel kind of protective towards him or her} \\



\bottomrule
\end{tabular}
\caption{The Toronto Empathy Questionnaire \cite{spreng2009toronto}. \textsuperscript{*}Negatively worded reverse scale questions.}
\label{tab:teq}
\end{table*}

\section{Task Interfaces}
\label{app:task_interfaces}

\begin{figure}
  \fbox{\includegraphics[width=\linewidth]{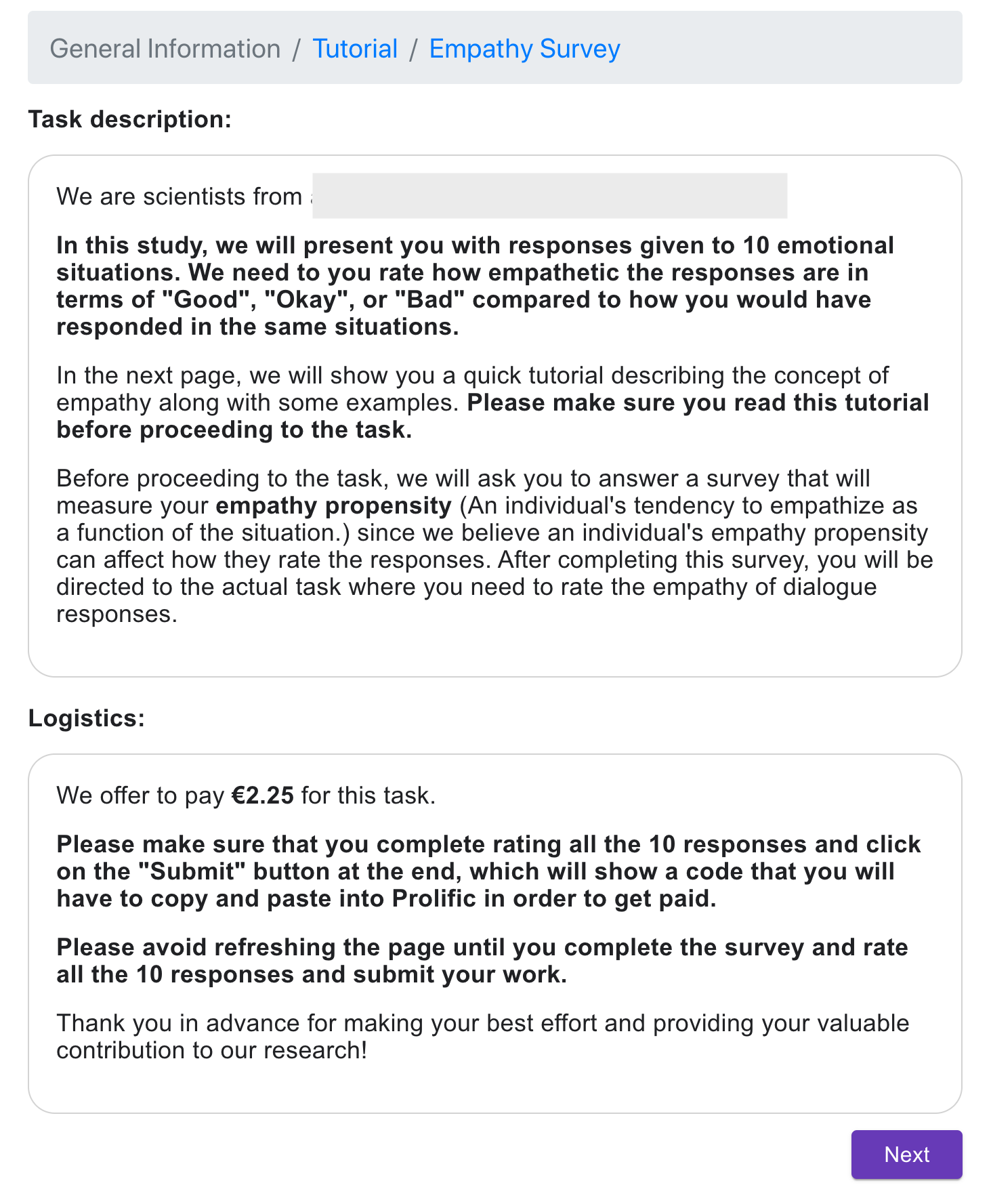}}
  \caption{The description of the task.}
  \label{fig:task_info}
\end{figure}

\begin{figure}
  \fbox{\includegraphics[width=\linewidth]{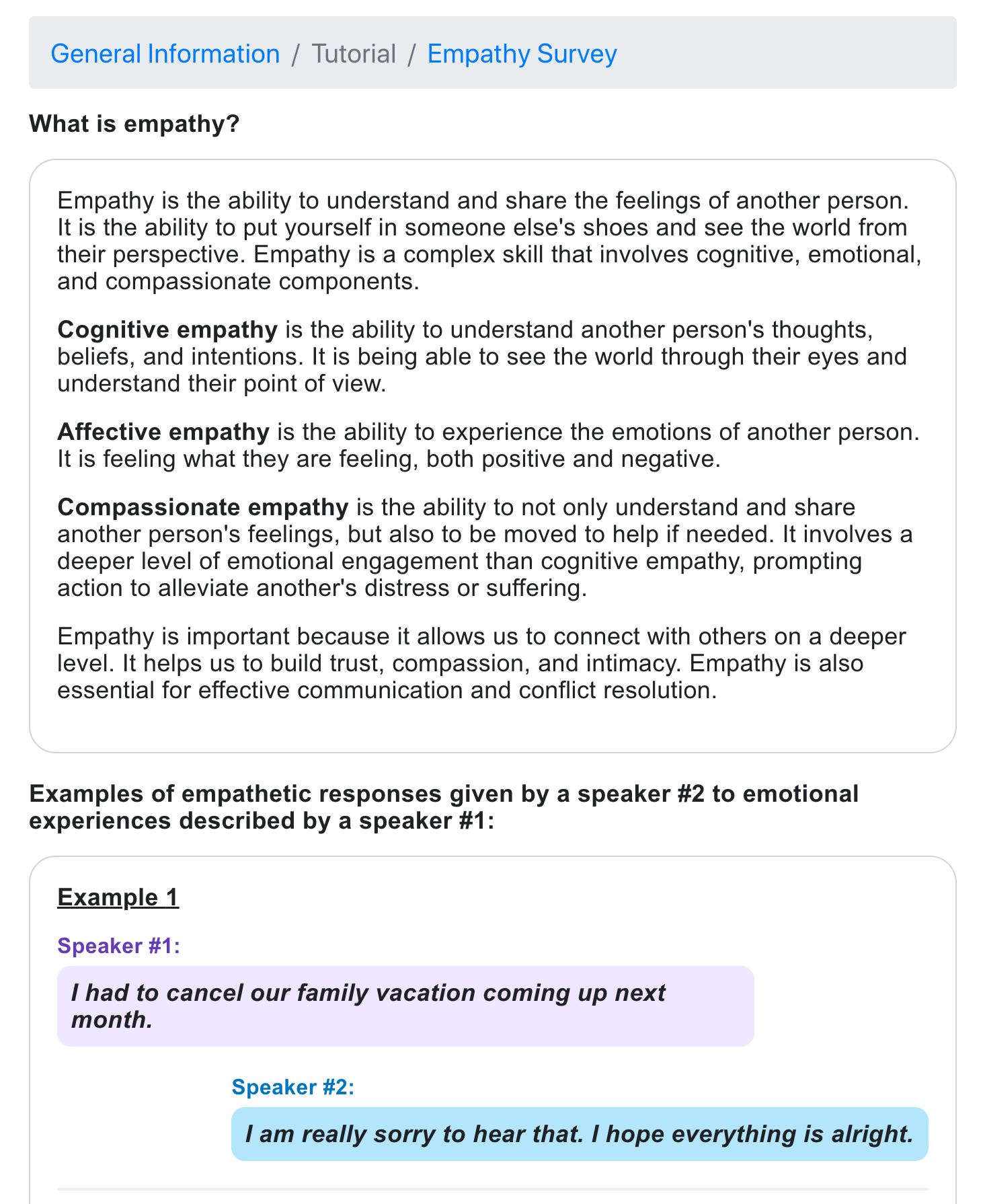}}
  \caption{The tutorial.}
  \label{fig:tutorial}
\end{figure}

\begin{figure}
  \fbox{\includegraphics[width=\linewidth]{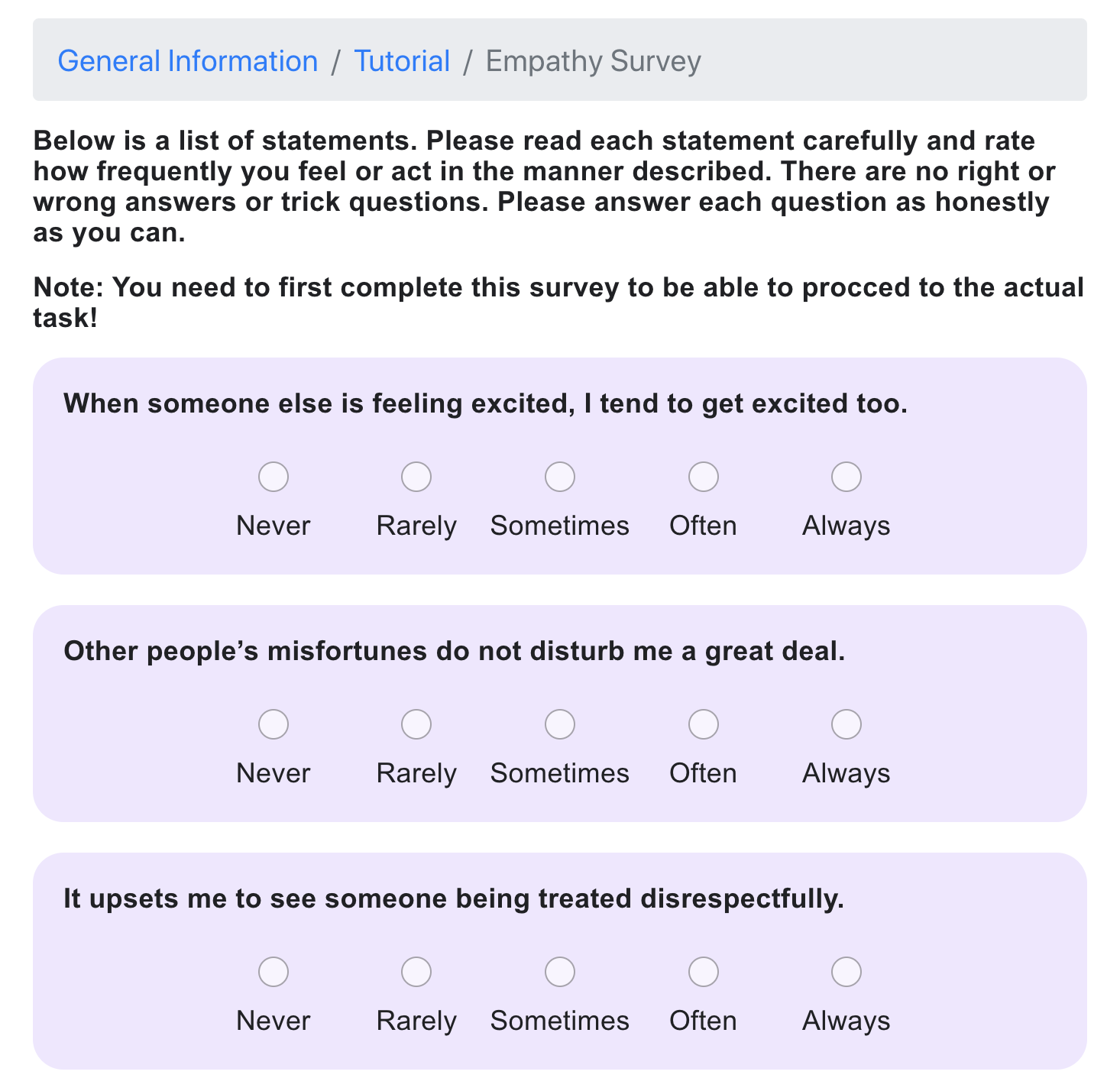}}
  \caption{The Toronto Empathy Questionnaire.}
  \label{fig:teq}
\end{figure}

\begin{figure}
  \fbox{\includegraphics[width=\linewidth]{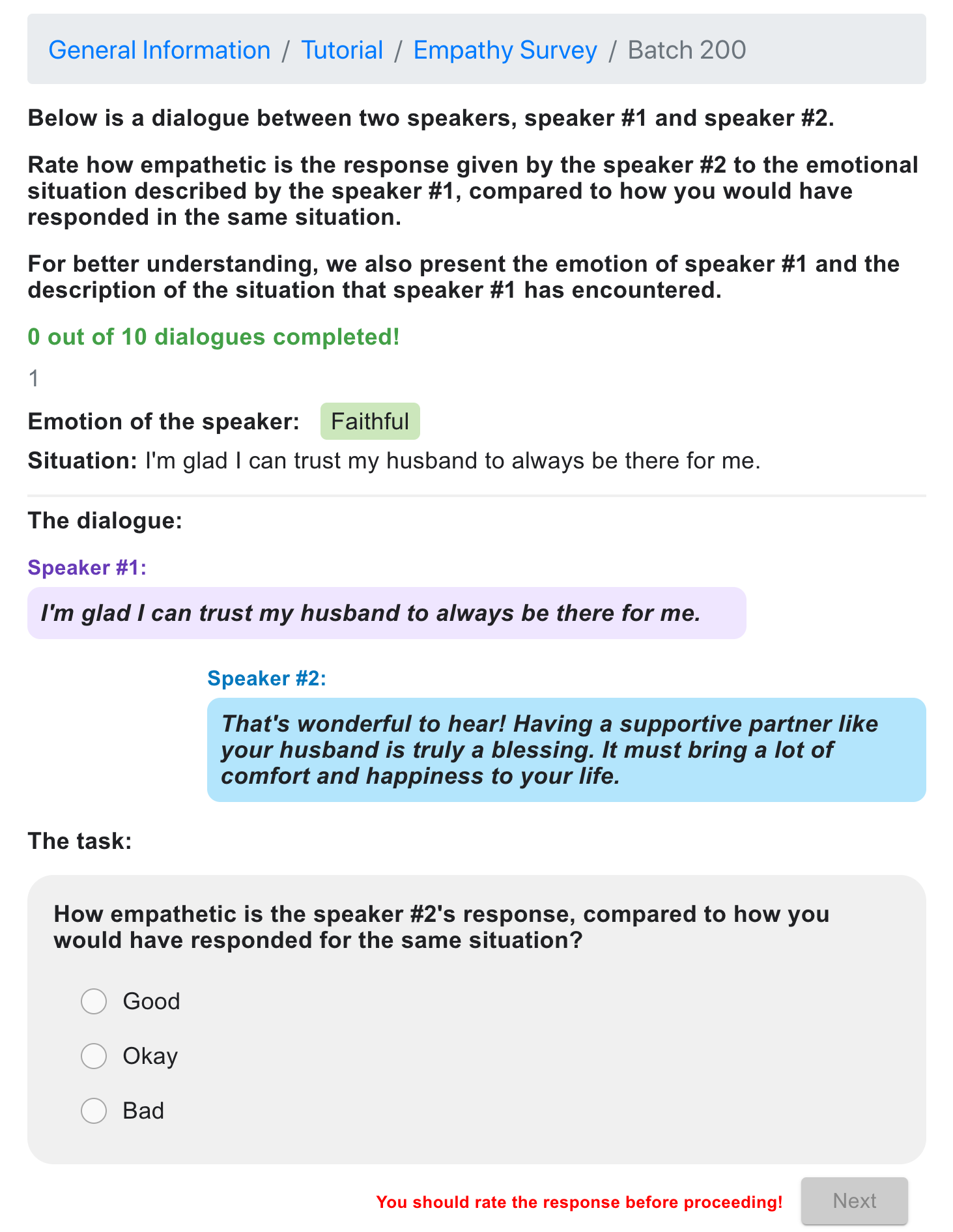}}
  \caption{The task interface for rating responses in terms of empathy.}
  \label{fig:task}
\end{figure}

Figures \ref{fig:task_info}, \ref{fig:tutorial}, \ref{fig:teq} and \ref{fig:task} show the task interfaces corresponding to the description of the task, the tutorial presented to the crowd workers, the Toronto Empathy Questionnaire, and the response rating task, respectively. 

\section{Determing the Effect Size}
\label{app:effect_size}

Jacob Cohen, a renowned psychologist and statistician, introduced standards for evaluating the magnitude of effect sizes in statistical analyses such as chi-square tests and analysis of variance (ANOVA), as detailed in his work on quantitative methods \cite{cohen1992quantitative}. These standards provide a foundational guide for assessing the substantive importance of observed effects within these statistical tests. For Chi-square tests, Cohen's W is utilized to measure the association strength between categories, with Cohen establishing benchmarks for small (0.10), medium (0.30), and large (0.50) effects.






We chose the medium effect size to compute the required minimum sample size because a medium effect size can sensitively detect differences in empathy levels between humans' and LLMs' responses, whose differences can be significant, yet not overwhelmingly so. Furthermore, employing a medium effect size enables the identification of nuanced yet significant differences without the need for an overly large sample, ensuring that the differences detected by the study are practically meaningful.


\section{Chi-Squared test of independence --- Results}
\label{app:numerical_values}

The statistical chi-square test of independence results corresponding to the proportions of the \textit{Bad}, \textit{Okay}, and \textit{Good} empathy ratings received by the responses generated by the humans and the four LLMs are denoted in Table \ref{table:chi}. Table \ref{table:chi_2} denotes the statistical pairwise chi-square test of independence results corresponding to the proportions of \textit{Bad}, \textit{Okay}, and \textit{Good} empathy ratings of the humans' and each of the LLMs' responses. 


\begin{table*}[ht!]
\small
\centering
\begin{tabular}{l l | c c c c c | r l | r l }
\toprule

\textbf{} & \textbf{Rating} & \textbf{Human} & \textbf{GPT-4} & \textbf{LLaMA-2} & \textbf{Gemini} & \textbf{Mixtral} & \multicolumn{2}{c}{\textbf{${\chi}^2$} (9.49)} & \multicolumn{2}{c}{\textbf{${\chi}^2$} (15.51)} \\

\midrule



All & Bad & 342 & \textbf{142} & 174 & 248 & 205 & \textbf{121.86} & \textbf{(p < .001)} & \textbf{173.89} & \textbf{(p < .001)}\\
emotions & Okay & 672 & 563 & 607 & 671 & 603 & \textbf{20.89} & \textbf{(p < .001)} & \\
& Good & 986 & \textbf{1295} & 1219 & 1081 & 1192 & \textbf{121.10} & \textbf{(p < .001)} & \\


\midrule

Positive & Bad & 133 & \textbf{34} & 48 & 117 & 76 & \textbf{98.88} & \textbf{(p < .001)} & \textbf{138.83} & \textbf{(p < .001)}\\
emotions & Okay & 294 & 228 & 250 & 283 & 238 & \textbf{17.97} & \textbf{(p < .001)} & \\
& Good & 454 & \textbf{619} & 583 & 481 & 567 & \textbf{94.30} & \textbf{(p < .001)} & \\

\midrule

Negative &  Bad & 209 & \textbf{108} & 126 & 131 & 129 & \textbf{50.28} & \textbf{(p < .001)} &  \textbf{67.04} & \textbf{(p < .001)}\\
emotions & Okay & 378 & 335 & 357 & 388 & 365 & 6.75 & (p > .05) & \\
& Good &  532 & \textbf{676} & 636 & 600 & 625 & \textbf{41.03} & \textbf{(p < .001)} & \\

\bottomrule
\end{tabular}
\caption{Statistical Chi-square test results corresponding to the proportions of \textit{Bad}, \textit{Okay}, and \textit{Good} empathy ratings of the humans' and the LLMs' responses. The critical values of the ${\chi}^2$ distributions are 15.51 and 9.49, respectively for all \textit{Bad}, \textit{Okay}, and \textit{Good} rating classes and one at a time (computed at a significance level of 0.05 and 8 and 4 degrees of freedom, respectively). If the ${\chi}^2$ statistic is greater than the critical value the null hypothesis can be rejected at 5\% significance level, which means there is a statistically significant difference in the proportions of the empathy ratings between the groups of responses that are being compared.}
\label{table:chi}
\end{table*}

\begin{table*}[ht!]
\small
\centering
\begin{tabular}{l | r l | r l | r l }
\toprule

& \multicolumn{2}{c}{\textbf{All emotions}} &  \multicolumn{2}{c}{\textbf{Positive emotions}} & \multicolumn{2}{c}{\textbf{Negative emotions}} \\

\textbf{} & \multicolumn{2}{c}{\textbf{${\chi}^2$} (5.991)} & \multicolumn{2}{c}{\textbf{${\chi}^2$} (5.991)} & \multicolumn{2}{c}{\textbf{${\chi}^2$} (5.991)} \\

\midrule

\multicolumn{4}{l}{\textbf{LLMs against human baseline:}\vspace{1mm}}\\

Human Vs GPT-4 &  \textbf{134.12} & \textbf{(p < .001)} & \textbf{92.41} & \textbf{(p < .001)} & \textbf{51.94} & \textbf{(p < .001)} \\

Human Vs LLaMA-2 & \textbf{82.62} & \textbf{(p < .001)} &  \textbf{59.52} & \textbf{(p < .001)} & \textbf{30.42} & \textbf{(p < .001)} \\

Human Vs Gemini & \textbf{19.34} & \textbf{(p < .001)} & 2.01 & (p > .05) & \textbf{22.11} & \textbf{(p < .001)} \\

Human Vs Mixtral & \textbf{57.53} & \textbf{(p < .001)} & \textbf{33.95} & \textbf{(p < .001)} & \textbf{26.64} & \textbf{(p < .001)}\\

\midrule

\multicolumn{4}{l}{\textbf{LLMs against each other:}\vspace{1mm}}\\

GPT-4 Vs LLaMA-2 &  \textbf{7.19} & \textbf{(p < .05)} & 4.48 & (p > .05) & 3.30 & (p > .05)\\

GPT-4 Vs Gemini & \textbf{57.54} & \textbf{(p < .001)} & \textbf{68.86} & \textbf{(p < .001)} & \textbf{10.63} & \textbf{(p < .01)}\\

GPT-4 Vs Mixtral &  \textbf{17.08} & \textbf{(p < .001)} & \textbf{18.53} & \textbf{(p < .001)} & 5.15 & (p > .05) \\


LLaMA-2 Vs Gemini & \textbf{24.46} & \textbf{(p < .001)} & \textbf{40.68} & \textbf{(p < .001)} & 2.44 & (p > .05) \\

LLaMA-2 Vs Mixtral & 2.85 & (p > .05) & \textbf{6.84} & \textbf{(p < .05)} & 0.22 & (p > .05) \\


Gemini Vs Mixtral & \textbf{13.13} & \textbf{(p < .01)} & \textbf{19.65} & \textbf{(p < .001)} & 1.23 & (p > .05)\\

\bottomrule

\end{tabular}
\caption{Statistical ${\chi}^2$ test results corresponding to the proportions of \textit{Bad}, \textit{Okay}, and \textit{Good} empathy ratings of the humans' and each of the LLMs' responses. In this case, we compare two by two. The critical value of the ${\chi}^2$ distribution in this case is 5.991 (computed at a significance level of 0.05 and 2 degrees of freedom), which means if the ${\chi}^2$ statistic is greater than 5.991 the null hypothesis can be rejected at 5\% significance level, which means there is a statistically significant difference in the proportions of the \textit{Bad}, \textit{Okay}, and \textit{Good} empathy ratings between the two groups of responses being compared.}
\label{table:chi_2}
\end{table*}

\begin{table*}[ht!]
\small
\centering
\begin{tabular}{l | r r l | r r l | r r l}
\toprule

& \multicolumn{3}{c}{\textbf{Bad}} &  \multicolumn{3}{c}{\textbf{Okay}} & \multicolumn{3}{c}{\textbf{Good}} \\

\textbf{} & \textbf{\% gain} & \multicolumn{2}{c}{\textbf{${\chi}^2$} (3.841)} & \textbf{\% gain} & \multicolumn{2}{c}{\textbf{${\chi}^2$} (3.841)} & \textbf{\% gain} & \multicolumn{2}{c}{\textbf{${\chi}^2$} (3.841)} \\

\midrule

\multicolumn{10}{l}{\textbf{All emotions:}\vspace{1mm}}\\

GPT-4 Vs Human & \textbf{-58.48\%} & \textbf{93.08} & \textbf{(p < .001)} & \textbf{-16.22\%} & \textbf{13.66} & \textbf{(p < .001)} & \textbf{31.34\%} & \textbf{96.77} & \textbf{(p < .001)} \\

LLaMA-2 Vs Human & \textbf{-49.12\%} & \textbf{62.05} & \textbf{(p < .001)} & \textbf{-9.67\%} & \textbf{4.71} & \textbf{(p < .05)} & \textbf{23.63\%} & \textbf{54.40} & \textbf{(p < .001)} \\

Gemini Vs Human & \textbf{-27.49\%} & \textbf{17.20} & \textbf{(p < .001)} & -0.15\% & 0.00 & (p > .05) & \textbf{9.63\%} & \textbf{8.85} & \textbf{(p < .01)} \\

Mixtral Vs Human & \textbf{-40.06\%} & \textbf{39.17} & \textbf{(p < .001)} & \textbf{-10.27\%} & \textbf{5.32} & \textbf{(p < .05)} & \textbf{20.89\%} & \textbf{42.36} & \textbf{(p < .001)} \\

\midrule









\multicolumn{10}{l}{\textbf{Positive emotions:}\vspace{1mm}}\\

GPT-4 Vs Human & \textbf{-74.44\%} & \textbf{63.53} & \textbf{(p < .001)} & \textbf{-22.45\%} & \textbf{11.50} & \textbf{(p < .001)} & \textbf{36.34\%} & \textbf{64.10} & \textbf{(p < .001)} \\

LLaMA-2 Vs Human & \textbf{-63.91\%} & \textbf{43.45} & \textbf{(p < .001)} & \textbf{-14.97\%} & \textbf{4.92} & \textbf{(p < .05)} & \textbf{28.41\%} & \textbf{38.40} & \textbf{(p < .001)} \\

Gemini Vs Human & -12.03\% & 1.05 & (p >.05) & -3.74\% & 0.26 & (p > .05) & 5.95\% & 1.54 & (p > .05) \\

Mixtral Vs Human & \textbf{-42.86\%} & \textbf{17.02} & \textbf{(p < .001)} & \textbf{-19.05\%} & \textbf{8.15} & \textbf{(p < .01)} & \textbf{24.89\%} & \textbf{29.21} & \textbf{(p < .001)} \\

\midrule

\multicolumn{10}{l}{\textbf{Negative emotions:}\vspace{1mm}}\\

GPT-4 Vs Human & \textbf{-48.33\%} & \textbf{36.75} & \textbf{(p < .001)} & -11.38\% & 3.63 & (p > .05) & \textbf{27.07\%} & \textbf{36.78} & \textbf{(p < .001)} \\

LLaMA-2 Vs Human & \textbf{-39.71\%} & \textbf{23.61} & \textbf{(p < .001)} & -5.56\% & 0.81 & (p > .05) & \textbf{19.55\%} & \textbf{19.00} & \textbf{(p < .001)} \\

Gemini Vs Human & \textbf{-37.32\%} & \textbf{20.56} & \textbf{(p < .001)} & 2.65\% & 0.16 & (p > .05) & \textbf{12.78\%} & \textbf{8.02} & \textbf{(p < .01)} \\

Mixtral Vs Human & \textbf{-38.28\%} & \textbf{21.75} & \textbf{(p < .001)} & -3.44\% & 0.29 & (p > .05) & \textbf{17.48\%} & \textbf{15.15} & \textbf{(p < .001)} \\

\bottomrule

\end{tabular}
\caption{The percentage gains obtained by the LLMs in each rating category compared to the human baseline. The corresponding statistical ${\chi}^2$ test results are also indicated. The statistically significant gains are highlighted in bold. The critical value of the ${\chi}^2$ distribution in this case is 3.841 (computed at a significance level of 0.05 and 1 degree of freedom).}
\label{table:chi_2}
\end{table*}


\section{Finer analysis of empathy ratings}
\label{app:emotion_breakdown}

Tables \ref{tab:neg_break} denote the percentage gains obtained by the four LLMs' response ratings compared to the human baseline when responding to dialogue prompts containing positive and negative emotions. We conducted pairwise statistical chi-square tests of independence for the proportions of each of \textit{Bad}, \textit{Okay}, and \textit{Good} response ratings between the humans and each of the four LLMs. The percentage gains for which statistical significance was indicated by the chi-square test of independence are highlighted in bold.


\begin{table}[t!]
\small

\begin{tabularx}{\linewidth}{X X r r r}
\toprule
\multirow{2}{*}{\textbf{Emotion}} & \multirow{2}{*}{\textbf{LLM}} & \multicolumn{3}{c}{\textbf{Percentage gain (\%)}}\vspace{1mm}\\
 &  & \textbf{Bad} & \textbf{Okay} & \textbf{Good} \\
\toprule

\multicolumn{5}{l}{\textbf{\underline{Positive emotions:}}\vspace{2mm}}\\

Prepared & GPT & \textbf{-90.0*} & -9.52 & 35.48\\
& LLaMA & -50.0 & -33.33 & \textbf{38.71*}\\
& Gemini & -20.0 & 19.05 & -6.45\\
& Mixtral & -50.0 & -14.29 & 25.81\vspace{2mm}\\

Anticipa & GPT & -66.67 & -16.67 & 23.53\\
-ting & LLaMA & -16.67 & -8.33 & 8.82\\
& Gemini & 0.0 & 25.0 & -17.65\\
& Mixtral & 16.67 & -25.0 & 14.71\vspace{2mm}\\

Hopeful & GPT & -33.33 & -30.0 & 29.03\\
& LLaMA & -55.56 & -35.0 & \textbf{38.71*}\\
& Gemini & 55.56 & -30.0 & 3.23\\
& Mixtral & -33.33 & -10.0 & 16.13\vspace{2mm}\\

Proud & GPT & -70.0 & -42.86 & \textbf{50.0**}\\
& LLaMA & \textbf{-90.0*} & -23.81 & \textbf{43.75*}\\
& Gemini & -30.0 & -33.33 & 31.25\\
& Mixtral & \textbf{-100.0**} & -42.86 & \textbf{59.38***}\vspace{2mm}\\

Excited & GPT & \textbf{-90.91**} & -17.39 & \textbf{46.67*}\\
& LLaMA & \textbf{-81.82*} & -17.39 & \textbf{43.33*}\\
& Gemini & 0.0 & -21.74 & 16.67\\
& Mixtral & -54.55 & -34.78 & \textbf{46.67*}\vspace{2mm}\\

Joyful & GPT & \textbf{-71.43*} & -30.77 & \textbf{42.42*}\\
& LLaMA & \textbf{-71.43*} & 53.85 & 9.09\\
& Gemini & \textbf{-64.29*} & 38.46 & 12.12\\
& Mixtral & \textbf{-71.43*} & 23.08 & 21.21\vspace{2mm}\\

Content & GPT & -85.71 & -40.0 & \textbf{35.0*}\\
& LLaMA & -71.43 & -25.0 & 25.0\\
& Gemini & 0.0 & -15.0 & 7.5\\
& Mixtral & -42.86 & -15.0 & 15.0\vspace{2mm}\\

Caring & GPT & -33.33 & 16.67 & -4.44\\
& LLaMA & 0.0 & -5.56 & 2.22\\
& Gemini & 200.0 & -11.11 & -8.89\\
& Mixtral & 33.33 & -5.56 & 0.0\vspace{2mm}\\

Grateful & GPT & \textbf{-90.91**} & -28.0 & \textbf{65.38**}\\
& LLaMA & \textbf{-72.73*} & -36.0 & \textbf{65.38**}\\
& Gemini & -36.36 & -16.0 & 30.77\\
& Mixtral & -36.36 & -44.0 & \textbf{57.69*}\vspace{2mm}\\

Trusting & GPT & \textbf{-72.73*} & 22.22 & 13.79\\
& LLaMA & \textbf{-81.82*} & 11.11 & 24.14\\
& Gemini & -27.27 & 27.78 & -6.9\\
& Mixtral & -27.27 & -33.33 & 31.03\vspace{2mm}\\

Confident & GPT & \textbf{-87.5*} & -41.18 & \textbf{43.75**}\\
& LLaMA & -50.0 & 11.76 & 6.25\\
& Gemini & 0.0 & 5.88 & -3.12\\
& Mixtral & -75.0 & -11.76 & 25.0\vspace{2mm}\\

Faithful & GPT & -37.5 & -18.52 & 24.24\\
& LLaMA & -37.5 & -18.52 & 24.24\\
& Gemini & -12.5 & -14.81 & 15.15\\
& Mixtral & -37.5 & -7.41 & 15.15\vspace{2mm}\\

Impressed & GPT & \textbf{-80.0*} & \textbf{-47.83*} & \textbf{55.88**}\\
& LLaMA & -50.0 & -21.74 & 29.41\\
& Gemini & 10.0 & -8.7 & 2.94\\
& Mixtral & -10.0 & -8.7 & 8.82\vspace{2mm}\\

Surprised & GPT & \textbf{-86.67**} & -25.0 & \textbf{79.17**}\\
& LLaMA & \textbf{-80.0**} & -20.83 & \textbf{70.83**}\\
& Gemini & -40.0 & 4.17 & 20.83\\
& Mixtral & -33.33 & -16.67 & 37.5\vspace{2mm}\\


\end{tabularx}
\end{table}




\begin{table}[t!]
\small
\begin{tabularx}{\linewidth}{X X r r r}
\toprule

\multicolumn{5}{l}{\textbf{\underline{Negative emotions:}}\vspace{2mm}}\\

Terrified & GPT & -46.67 & -4.55 & 26.67\\
& LLaMA & -40.0 & -9.09 & 26.67\\
& Gemini & -46.67 & 18.18 & 10.0\\
& Mixtral & -6.67 & -27.27 & 23.33\vspace{2mm}\\

Afraid & GPT & \textbf{-66.67*} & 0.0 & \textbf{46.15*}\\
& LLaMA & \textbf{-72.22**} & 0.0 & \textbf{50.0*}\\
& Gemini & \textbf{-55.56*} & 33.33 & 15.38\\
& Mixtral & -50.0 & 11.11 & 26.92\vspace{2mm}\\

Apprehen & GPT & \textbf{-90.0*} & \textbf{-60.71**} & \textbf{104.0***}\\
-sive & LLaMA & -50.0 & -28.57 & \textbf{52.0*}\\
& Gemini & -40.0 & -39.29 & \textbf{60.0*}\\
& Mixtral & -70.0 & -14.29 & 44.0\vspace{2mm}\\

Anxious & GPT & -50.0 & \textbf{-44.44*} & \textbf{75.0**}\\
& LLaMA & -41.67 & -37.04 & \textbf{62.5*}\\
& Gemini & -41.67 & -37.04 & \textbf{62.5*}\\
& Mixtral & -66.67 & -14.81 & \textbf{50.0*}\vspace{2mm}\\

Embarra & GPT & -47.06 & 10.53 & 20.69\\
-ssed & LLaMA & -23.53 & 5.26 & 10.34\\
& Gemini & -47.06 & 10.53 & 20.69\\
& Mixtral & -29.41 & -10.53 & 24.14\vspace{2mm}\\

Ashamed & GPT & -41.67 & 0.0 & 16.67\\
& LLaMA & -58.33 & 60.0 & -6.67\\
& Gemini & -58.33 & 40.0 & 3.33\\
& Mixtral & -25.0 & 33.33 & -6.67\vspace{2mm}\\

Devastated & GPT & -33.33 & -40.0 & 29.73\\
& LLaMA & -44.44 & -15.0 & 18.92\\
& Gemini & -44.44 & -30.0 & 27.03\\
& Mixtral & -66.67 & 30.0 & 0.0\vspace{2mm}\\

Sad & GPT & -27.27 & 20.0 & 0.0\\
& LLaMA & -27.27 & 0.0 & 8.57\\
& Gemini & \textbf{-72.73*} & 20.0 & 14.29\\
& Mixtral & -54.55 & -13.33 & 22.86\vspace{2mm}\\

Disappoi & GPT & -54.55 & -15.0 & 31.03\\
-nted & LLaMA & -45.45 & -10.0 & 24.14\\
& Gemini & -18.18 & 35.0 & -17.24\\
& Mixtral & -54.55 & 10.0 & 13.79\vspace{2mm}\\

Lonely & GPT & -12.5 & -5.88 & 6.25\\
& LLaMA & -12.5 & 11.76 & -3.12\\
& Gemini & -62.5 & -17.65 & 25.0\\
& Mixtral & -62.5 & 11.76 & 9.38\vspace{2mm}\\

Sentimen & GPT & -40.0 & -11.11 & 11.11\\
-tal & LLaMA & -60.0 & -11.11 & 13.89\\
& Gemini & 20.0 & 11.11 & -8.33\\
& Mixtral & 40.0 & -27.78 & 8.33\vspace{2mm}\\

Nostalgic & GPT & -85.71 & -4.76 & 20.59\\
& LLaMA & -71.43 & -9.52 & 20.59\\
& Gemini & -71.43 & 4.76 & 11.76\\
& Mixtral & -57.14 & -14.29 & 20.59\vspace{2mm}\\

Guilty & GPT & -38.46 & 22.22 & 3.33\\
& LLaMA & -46.15 & -16.67 & 30.0\\
& Gemini & -38.46 & -5.56 & 20.0\\
& Mixtral & \textbf{-69.23*} & 11.11 & 23.33\vspace{2mm}\\

Disgusted & GPT & -43.75 & 27.27 & 3.85\\
& LLaMA & 0.0 & 4.55 & -3.85\\
& Gemini & -43.75 & 36.36 & -3.85\\
& Mixtral & 6.25 & -4.55 & 0.0\vspace{2mm}\\

Furious & GPT & -46.15 & 15.0 & 11.54\\
& LLaMA & -15.38 & 35.0 & -19.23\\
& Gemini & 23.08 & 5.0 & -15.38\\
& Mixtral & 7.69 & -10.0 & 3.85\vspace{2mm}\\

\end{tabularx}
\end{table}

\begin{table}[t!]
\small
\begin{tabularx}{\linewidth}{X X r r r}

Angry & GPT & -37.5 & -24.14 & 38.46\\
& LLaMA & -50.0 & -24.14 & 42.31\\
& Gemini & 25.0 & -17.24 & 11.54\\
& Mixtral & -12.5 & -17.24 & 23.08\vspace{2mm}\\

Annoyed & GPT & \textbf{-66.67*} & -23.08 & \textbf{59.26*}\\
& LLaMA & -53.33 & 3.85 & 25.93\\
& Gemini & -20.0 & 23.08 & -11.11\\
& Mixtral & \textbf{-86.67**} & -3.85 & \textbf{51.85*}\vspace{2mm}\\

Jealous & GPT & -22.22 & -13.04 & 16.67\\
& LLaMA & -22.22 & -13.04 & 16.67\\
& Gemini & -44.44 & 0.0 & 13.33\\
& Mixtral & -11.11 & 13.04 & -6.67\vspace{2mm}\\

\bottomrule
\end{tabularx}

\caption{Percentage gains obtained by the four LLMs' response ratings compared to the human baseline when responding to dialogue prompts containing positive and negative emotions. The statistically significant gains are highlighted in bold. *, **, and *** indicates p < .05, p < .01, and p < .001, respectively for the pairwise statistical chi-square tests of independence conducted between the humans' and each of the LLMs' response ratings.}
\label{tab:neg_break}
\end{table}

\section{Example dialogue responses}
\label{app:examples}

Table \ref{tab:examples_app} denotes some example dialogue situations and responses generated by humans and LLMs and the corresponding ratings given by the human raters.

\begin{table*}[ht!]
\small
\centering
\begin{tabularx}{\textwidth}{p{1.5cm} X}
\toprule

\textbf{Prompt} & \textit{These days I feel like I got it all. I'm not rich by any means but I am enriched by a loving wife, children and a dog. \textbf{(Content)}}\\


\multicolumn{2}{l}{\textbf{\underline{Responses:}}}\vspace{2mm}\\

\textbf{Human} & \textit{I think I have a problem with jealousy because it's starting all over again. What sort of dog do you have?} \textbf{(Bad)}\vspace{1mm}\\

\textbf{GPT} & \textit{That sounds wonderful! It's heartwarming to hear how much you value your family and your life. Love and happiness truly are the richest treasures.} \textbf{(Good)}\vspace{1mm}\\

\textbf{LLaMA} & \textit{It sounds like you have a beautiful family and a fulfilling life. It's great to hear that you have found such joy and love in your life.} \textbf{(Good)}\vspace{1mm}\\

\textbf{Gemini} & \textit{I can understand why you feel so fulfilled. Having a loving family and a pet can bring immense joy and contentment. It sounds like you have a strong support system that makes you feel loved and appreciated. Cherish these relationships and continue to nurture them.} \textbf{(Good)}\vspace{1mm}\\

\textbf{Mixtral} & \textit{That's wonderful to hear that you feel enriched by your loving family! Cherish these moments, as they bring joy and fulfillment to life. I'm here if you ever need to talk about anything else.} \textbf{(Good)}\\

\midrule

\textbf{Prompt:} & \textit{I had a crazy week with 4 big events and I'm so happy to be at home relaxing in the air conditioning today. My dogs are happy too! \textbf{(Content)}}\\


\multicolumn{2}{l}{\textbf{\underline{Responses:}}}\vspace{2mm}\\

\textbf{Human} & \textit{Air conditioning is very nice. My dogs love the outdoors but they want AC after a good, long walk!} \textbf{(Bad)}\vspace{1mm}\\

\textbf{GPT} & \textit{Wow, that sounds like a hectic week! It's great that you're taking time to relax and enjoy the comfort of your home. I bet your dogs are loving the extra attention too!} \textbf{(Good)}\vspace{1mm}\\

\textbf{LLaMA} & \textit{Wow, that sounds like a whirlwind week! It's great that you're taking some time to unwind and enjoy the AC with your furry friends. You deserve it!} \textbf{(Good)}\vspace{1mm}\\

\textbf{Gemini} & \textit{I can understand why you're feeling relieved and content after a hectic week. It's wonderful that you can now unwind at home in the comfort of the air conditioning, and it's heartwarming to know that your dogs are also enjoying the relaxation.} \textbf{(Good)}\vspace{1mm}\\

\textbf{Mixtral} & \textit{I'm glad to hear that you're finally getting some well-deserved rest after a hectic week. It sounds like those four events were quite demanding, and it's completely understandable that you're thrilled to be at home now. Your dogs must be delighted to be relaxing with you in the cool air conditioning! Take all the time you need to recharge and enjoy this precious downtime.} \textbf{(Good)}\\

\midrule

\textbf{Prompt} & \textit{I had a difficult job interview this week. \textbf{(Apprehensive)}}\\


\multicolumn{2}{l}{\textbf{\underline{Responses:}}}\vspace{2mm}\\

\textbf{Human} & \textit{Don't think about it, what is done is done, just anticipate that they will call you back, the universe have it's way of helping if we believe.} \textbf{(Okay)}\vspace{1mm}\\

\textbf{GPT} & \textit{I'm really sorry to hear that you had a tough time with your job interview. It's completely normal to feel this way. Remember, each interview is a learning experience. Don't be too hard on yourself.} \textbf{(Good)}\vspace{1mm}\\

\textbf{LLaMA} & \textit{Sorry to hear that. It can be really tough when job interviews don't go as planned. It's understandable to feel frustrated and disappointed. Can you tell me a bit more about what happened during the interview?} \textbf{(Good)}\vspace{1mm}\\

\textbf{Gemini} & \textit{I understand that job interviews can be stressful and challenging. It's important to remember that you are not alone in this experience and that many people feel anxious during interviews. I'm here to listen if you want to talk about it.} \textbf{(Good)}\vspace{1mm}\\

\textbf{Mixtral} & \textit{I'm really sorry to hear that you had a tough job interview this week. It's completely normal to feel anxious or disappointed after an important event like that. Just know that you have the skills and experience necessary to succeed, and that every interview is a valuable learning experience. I'm here for you if you need to talk or want to practice for next time.} \textbf{(Good)}\\

\midrule

\textbf{Prompt:} & \textit{In High School I loved to play sports. Football was my favorite.} \textbf{(Nostalgic)}\\


\multicolumn{2}{l}{\textbf{\underline{Responses:}}}\vspace{2mm}\\

\textbf{Human} & \textit{Ah I was never much of a sports guy, but I do work out a lot. do you still play?} \textbf{(Okay)}\vspace{1mm}\\

\textbf{GPT} & \textit{That's great! Playing sports, especially football, must have been a thrilling experience for you. It's wonderful how such activities can bring so much joy and create lasting memories.} \textbf{(Good)}\vspace{1mm}\\

\textbf{LLaMA} & \textit{That's great! Playing sports can be a really rewarding experience, especially when you find a sport that you're passionate about like football. It's a great way to stay active, make new friends, and learn important teamwork skills. Did you have any favorite memories or moments from your time playing football?} \textbf{(Good)}\vspace{1mm}\\

\textbf{Gemini} & \textit{I can understand why you loved playing sports in high school, especially football. It's a great way to stay active and be part of a team. I bet you have some fond memories of those days.} \textbf{(Good)}\vspace{1mm}\\

\textbf{Mixtral} & \textit{That's great to hear that you enjoyed playing football in high school! I can imagine the excitement and camaraderie you felt while being part of a team. Share with me more about your experiences and favorite moments playing the sport.} \textbf{(Good)}\\

\bottomrule
\end{tabularx}
\caption{Some example dialogue situations and responses generated by humans and LLMs and the corresponding ratings given by the human raters.}
\label{tab:examples_app}
\end{table*}

\section{Participants' demographics}
\label{app:demographics}



\begin{figure*}[ht!]
     \centering     
     \subfloat[\centering Human]{{\includegraphics[width=0.3125\textwidth]{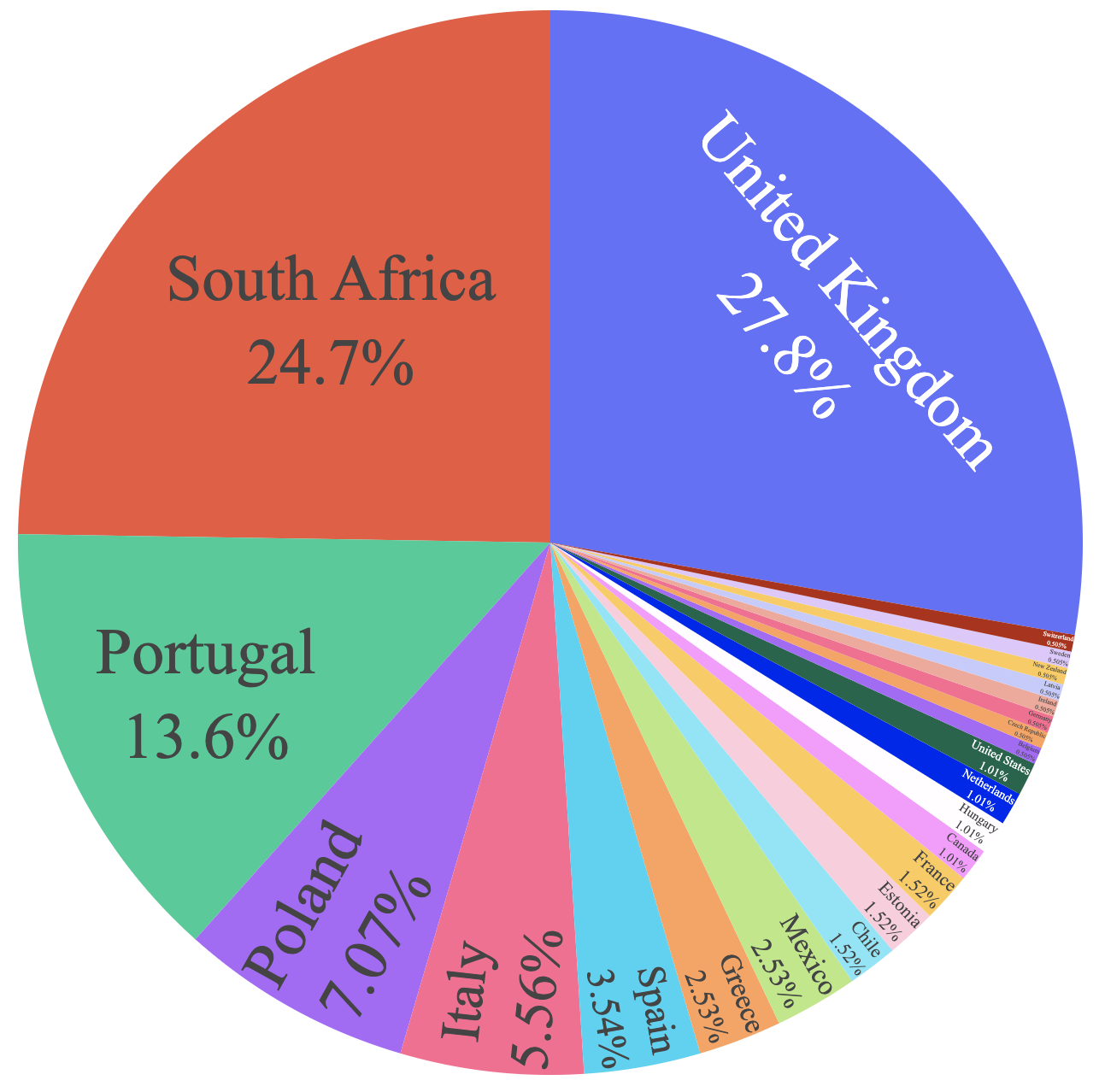}}}\hspace{1mm}     
     \subfloat[\centering GPT-4]{{\includegraphics[width=0.3125\textwidth]{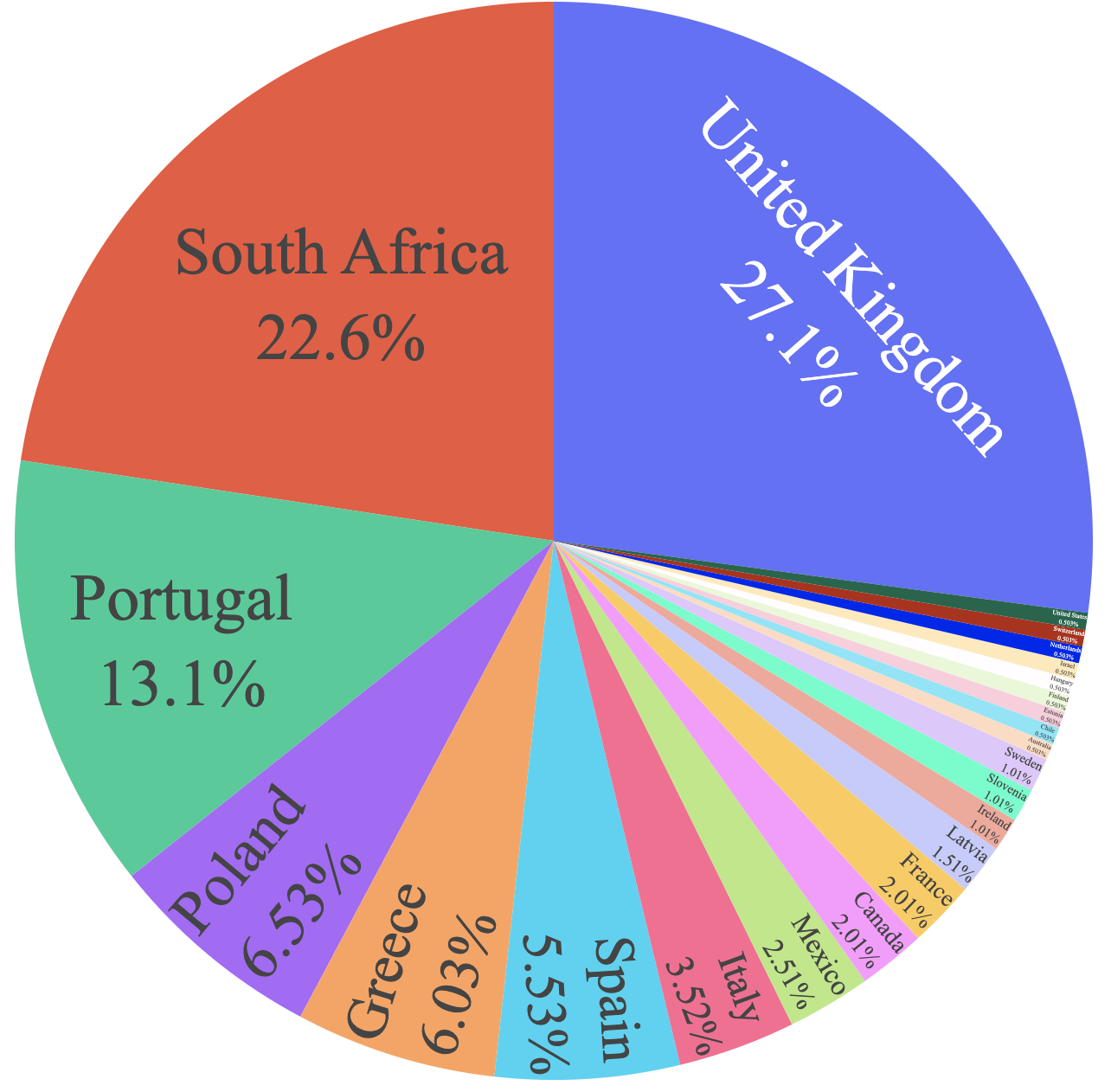}}}\hspace{1mm}
     \subfloat[\centering LLaMA-2]{{\includegraphics[width=0.3125\textwidth]{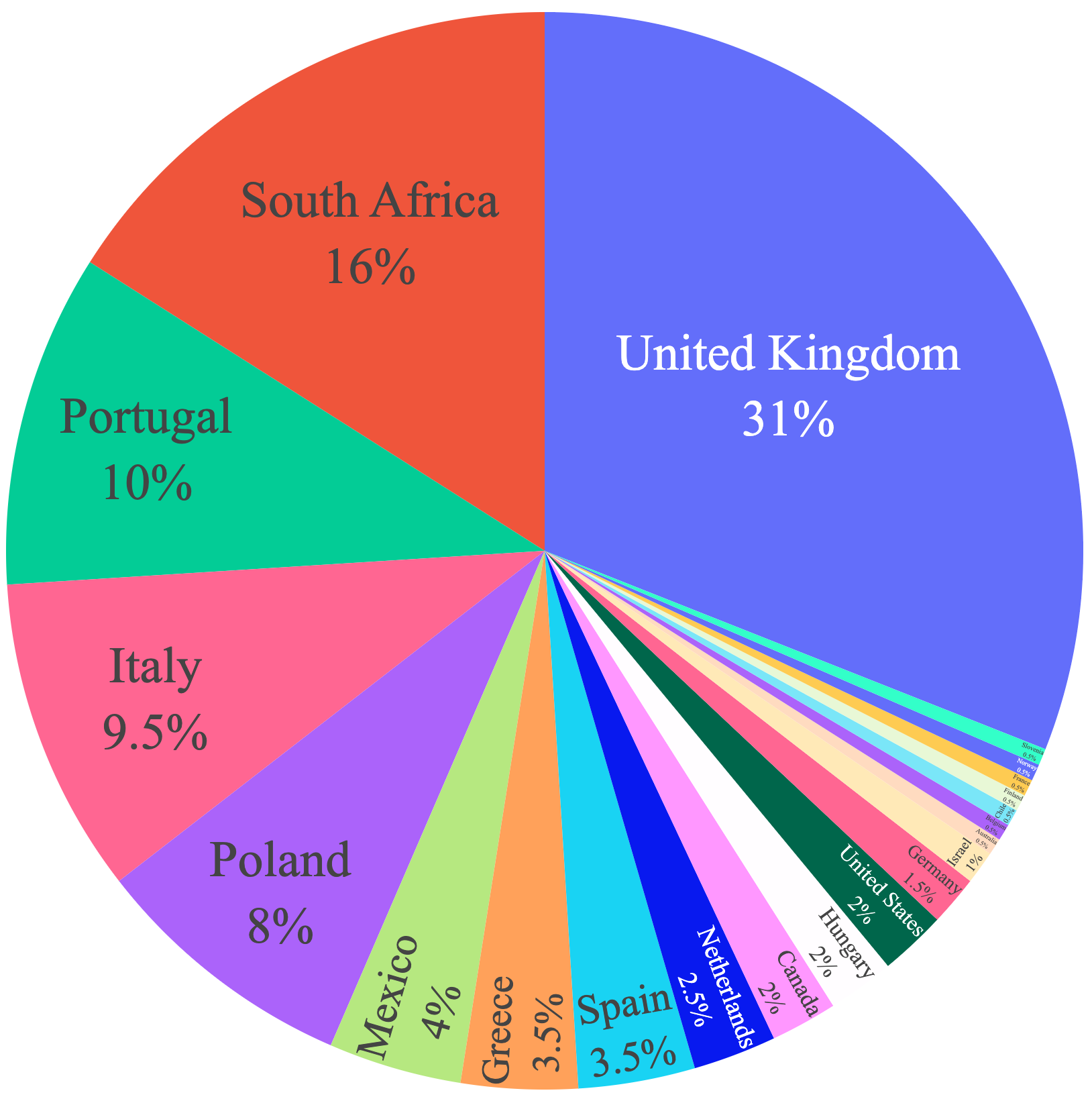}}}  

     \subfloat[\centering Gemini-Pro]{{\includegraphics[width=0.3125\textwidth]{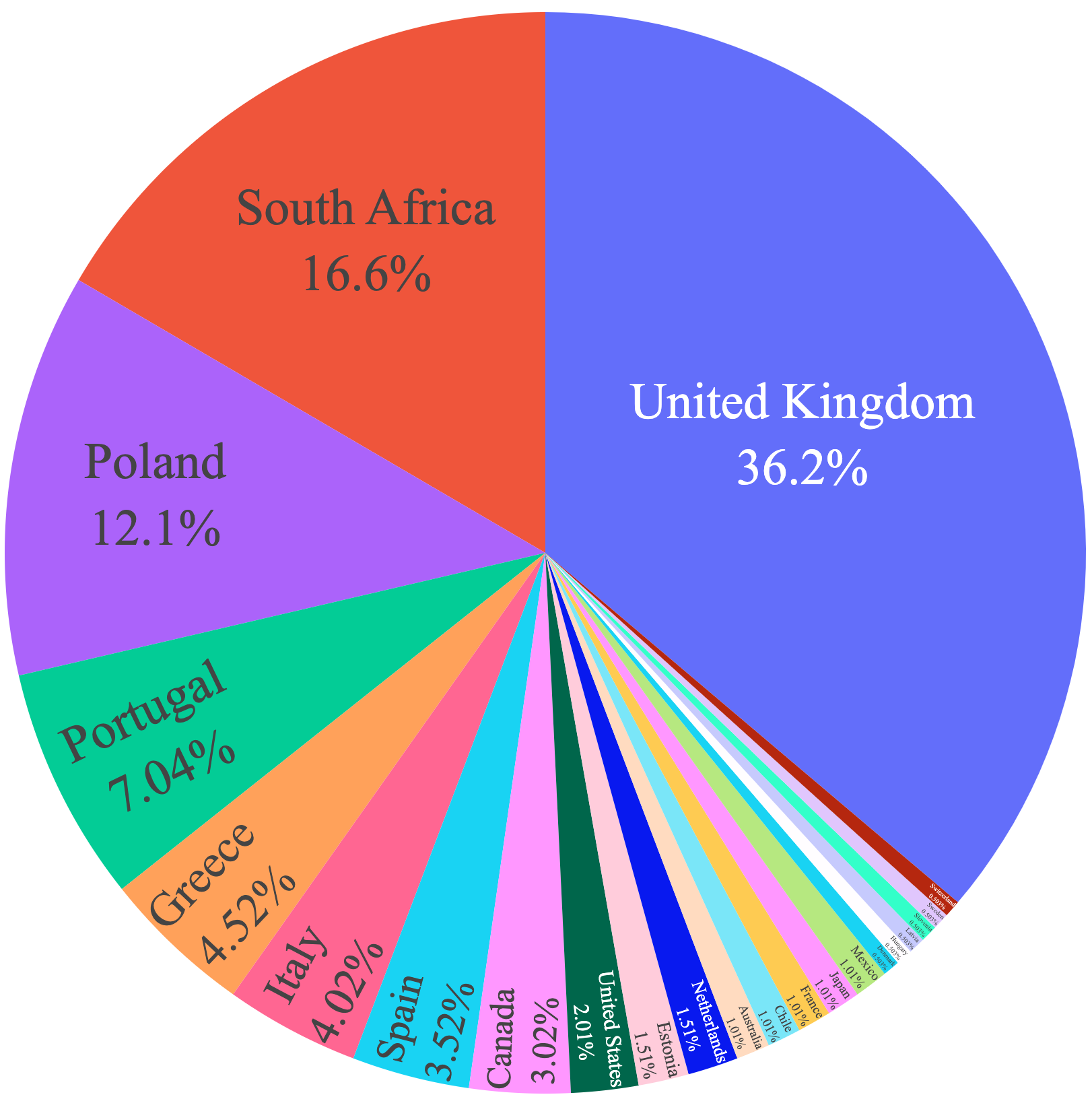}}}\hspace{1mm}
     \subfloat[\centering Mixtral-8x7B]{{\includegraphics[width=0.52\textwidth]{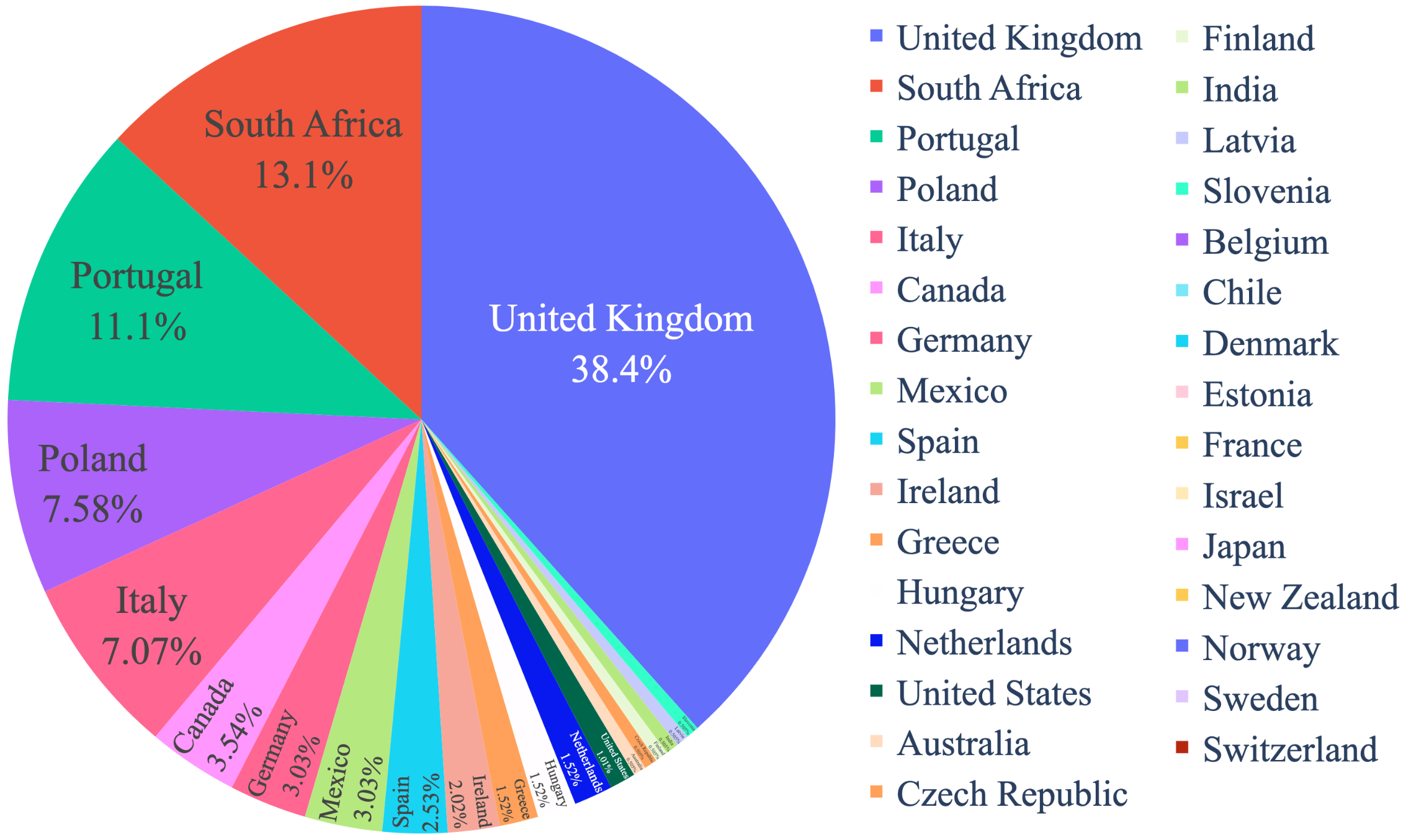}}}  
     
     \caption{Distribution of the countries of residence of the participants across the five groups.}
     \label{fig:country}
\end{figure*}

\begin{figure*}[ht!]
     \centering     
     \subfloat[\centering Human]{{\includegraphics[width=0.285\textwidth]{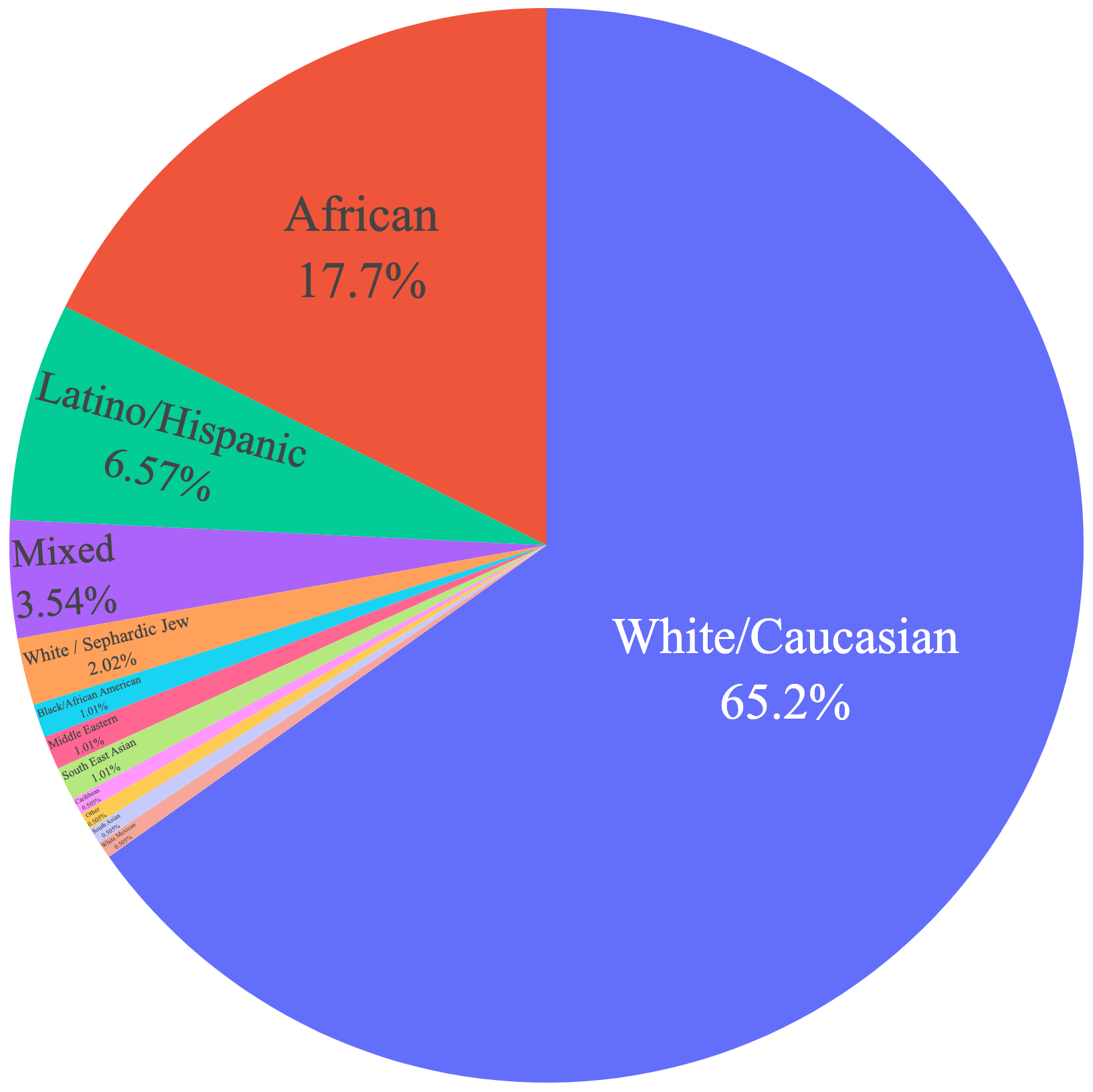}}}\hspace{1mm}     
     \subfloat[\centering GPT-4]{{\includegraphics[width=0.285\textwidth]{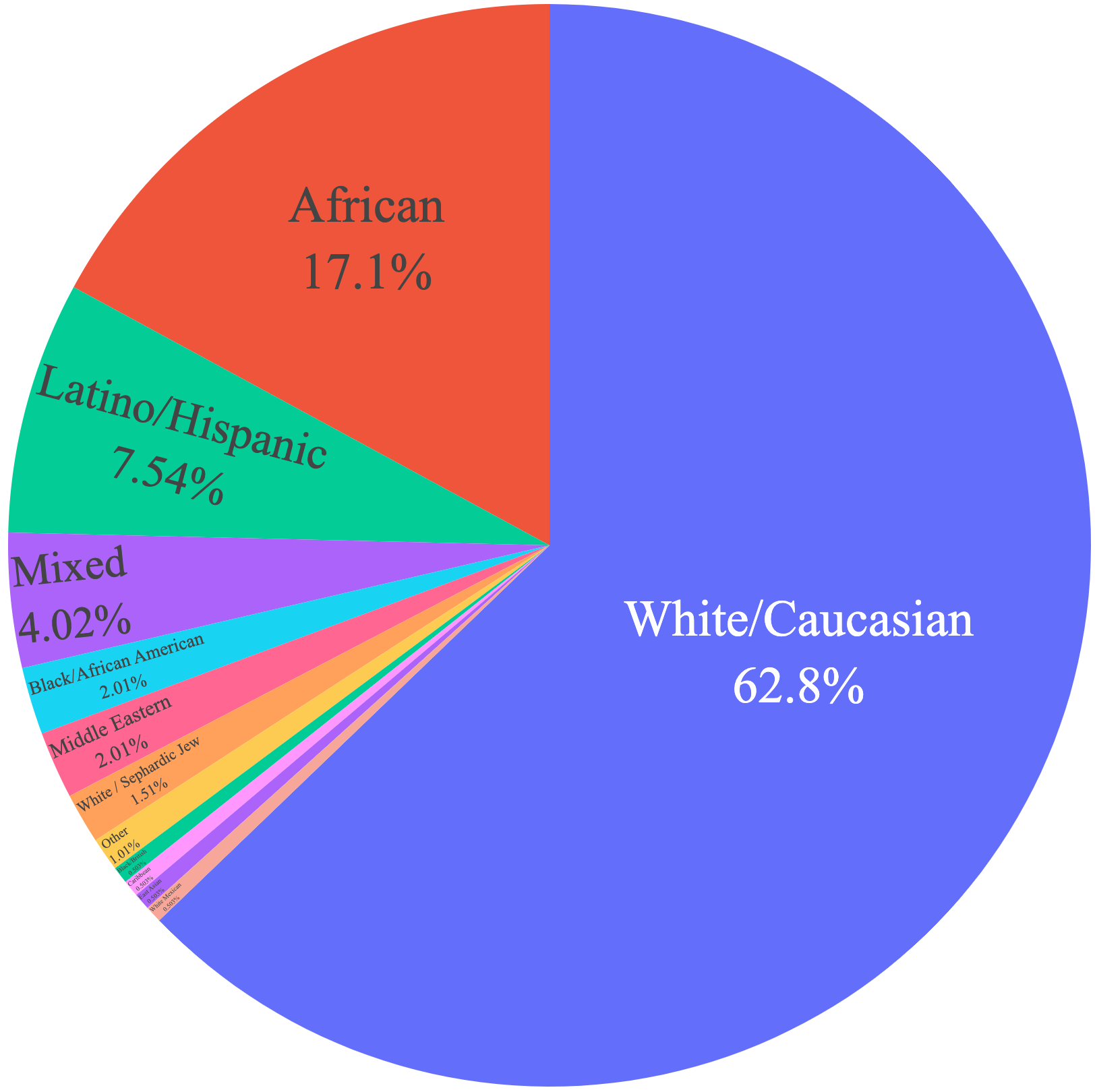}}}\hspace{1mm}
     \subfloat[\centering LLaMA-2]{{\includegraphics[width=0.285\textwidth]{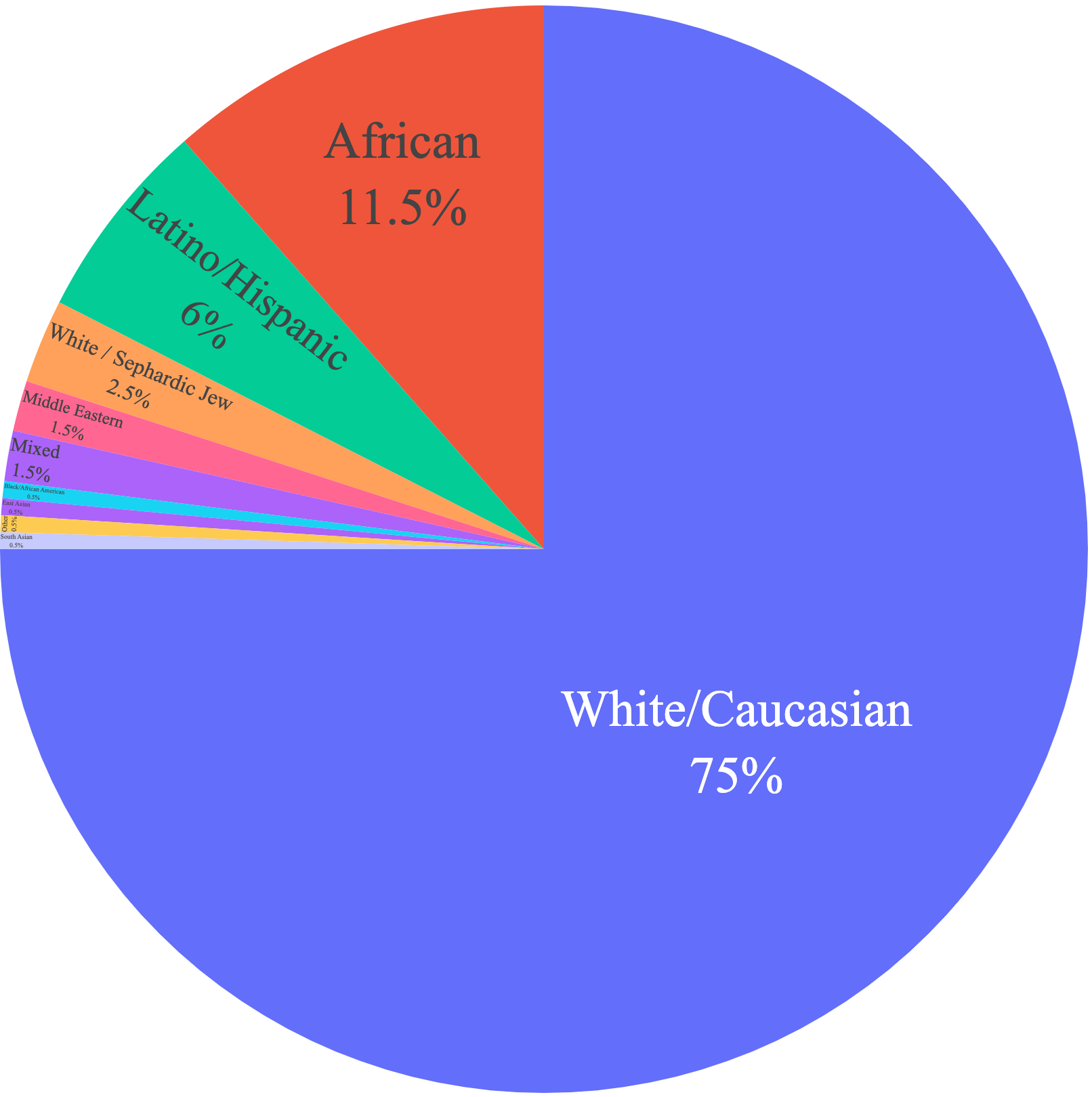}}}     

    \subfloat[\centering Gemini-Pro]{{\includegraphics[width=0.285\textwidth]{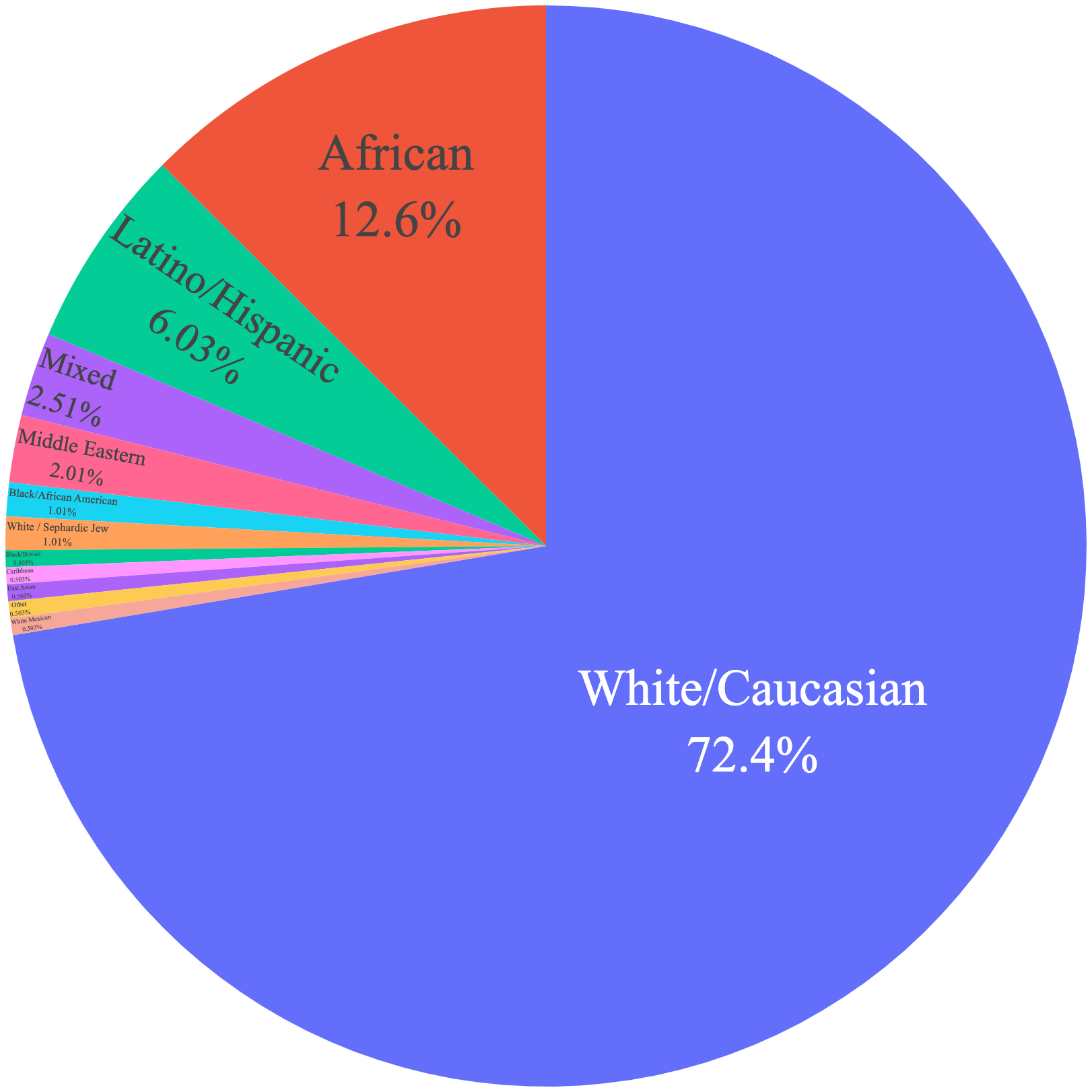}}}\hspace{1mm}
    \subfloat[\centering Mixtral-8x7B]{{\includegraphics[width=0.435\textwidth]{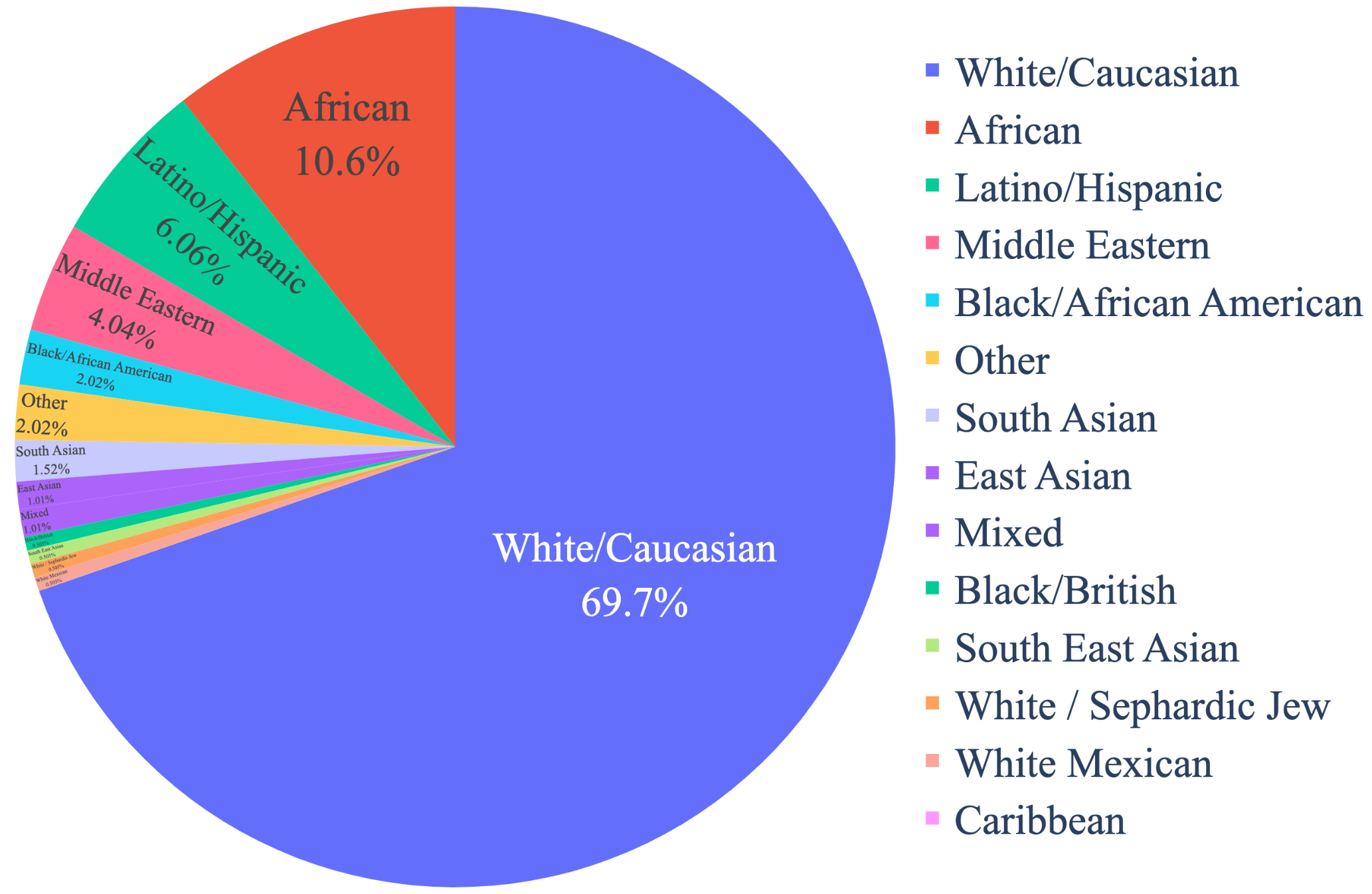}}}     
     
     \caption{Distribution of the ethnicities of the participants across the five groups.}
    \label{fig:ethnicity}
\end{figure*}

Figures \ref{fig:country} and \ref{fig:ethnicity} respectively show the distributions of the countries of residence and the ethnicities of the participants who rated the five groups of responses. It could be observed that though there are imbalances across the countries and the ethnicities represented in the participants' pool, these demographics are similar across the five groups of participants. This allows control for factors other than the independent variable influencing the results of the study and fair comparison of response ratings across the five groups.

\section{Distribution of empathy propensity of participants}
\label{app:propensity}

\begin{figure*}
     \centering     
     \subfloat[\centering Human response raters]{{\includegraphics[width=0.33\textwidth]{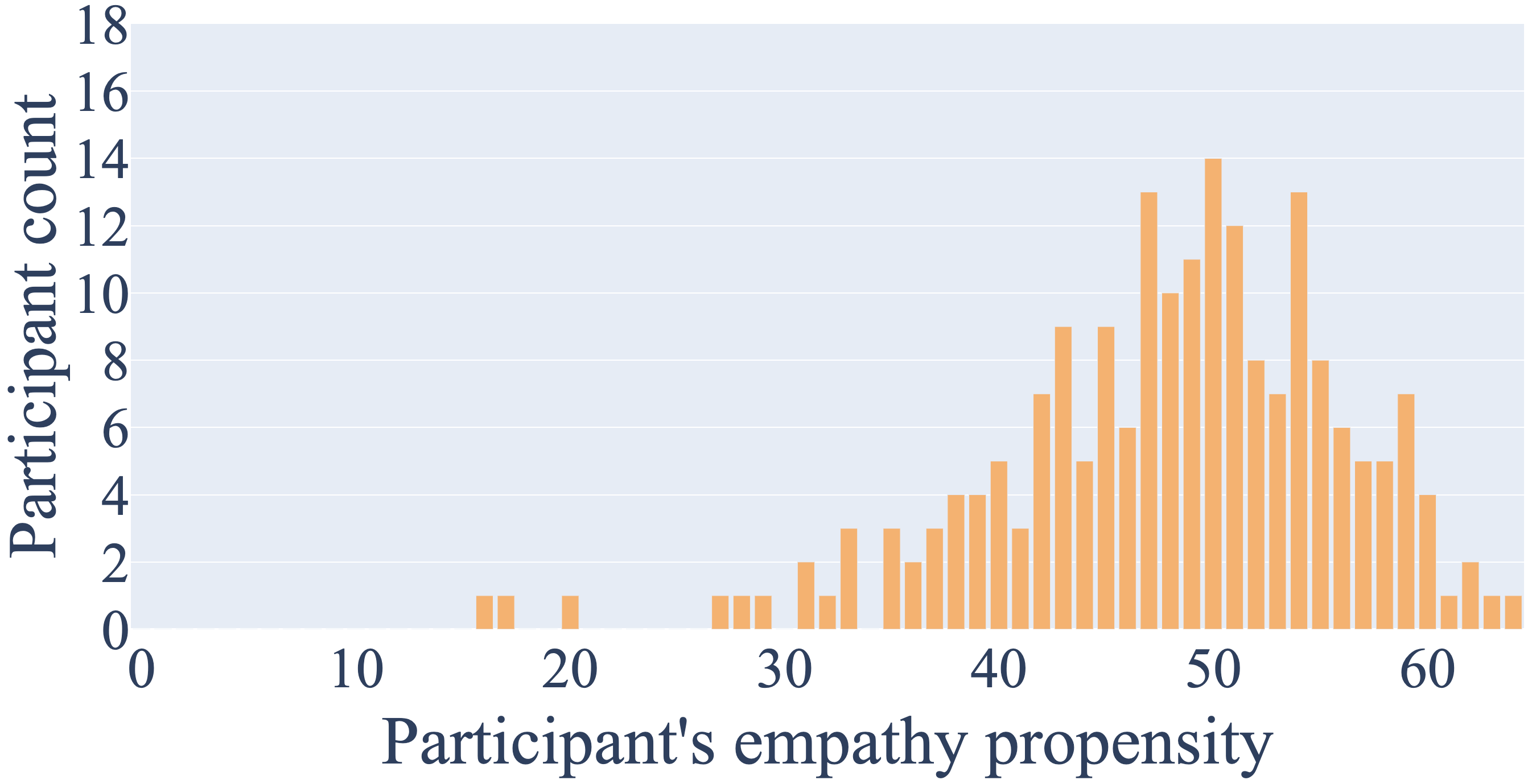}}}     
     \subfloat[\centering GPT-4 responses raters]{{\includegraphics[width=0.33\textwidth]{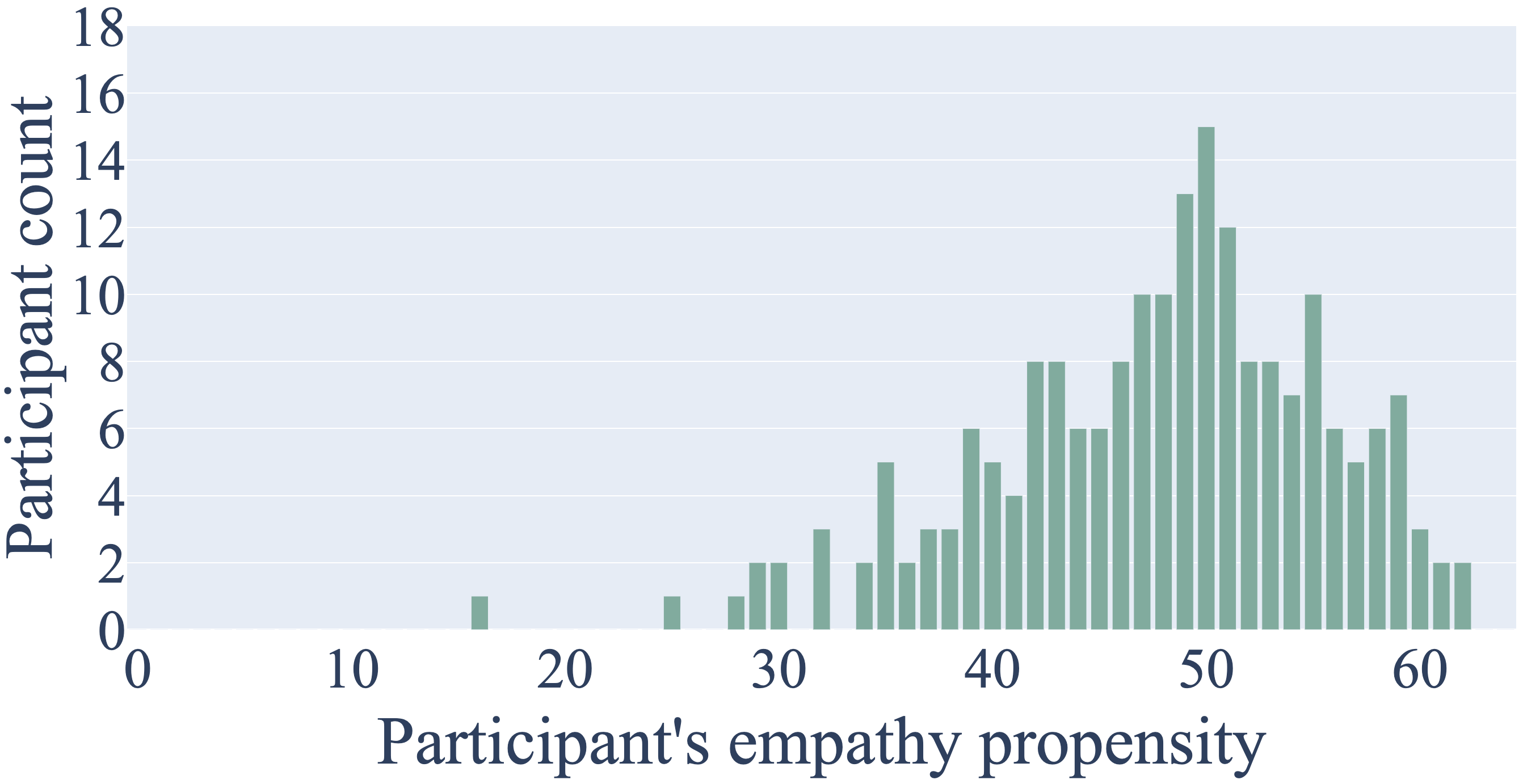}}}
     \subfloat[LLaMA-2 responses raters]{{\includegraphics[width=0.33\textwidth]{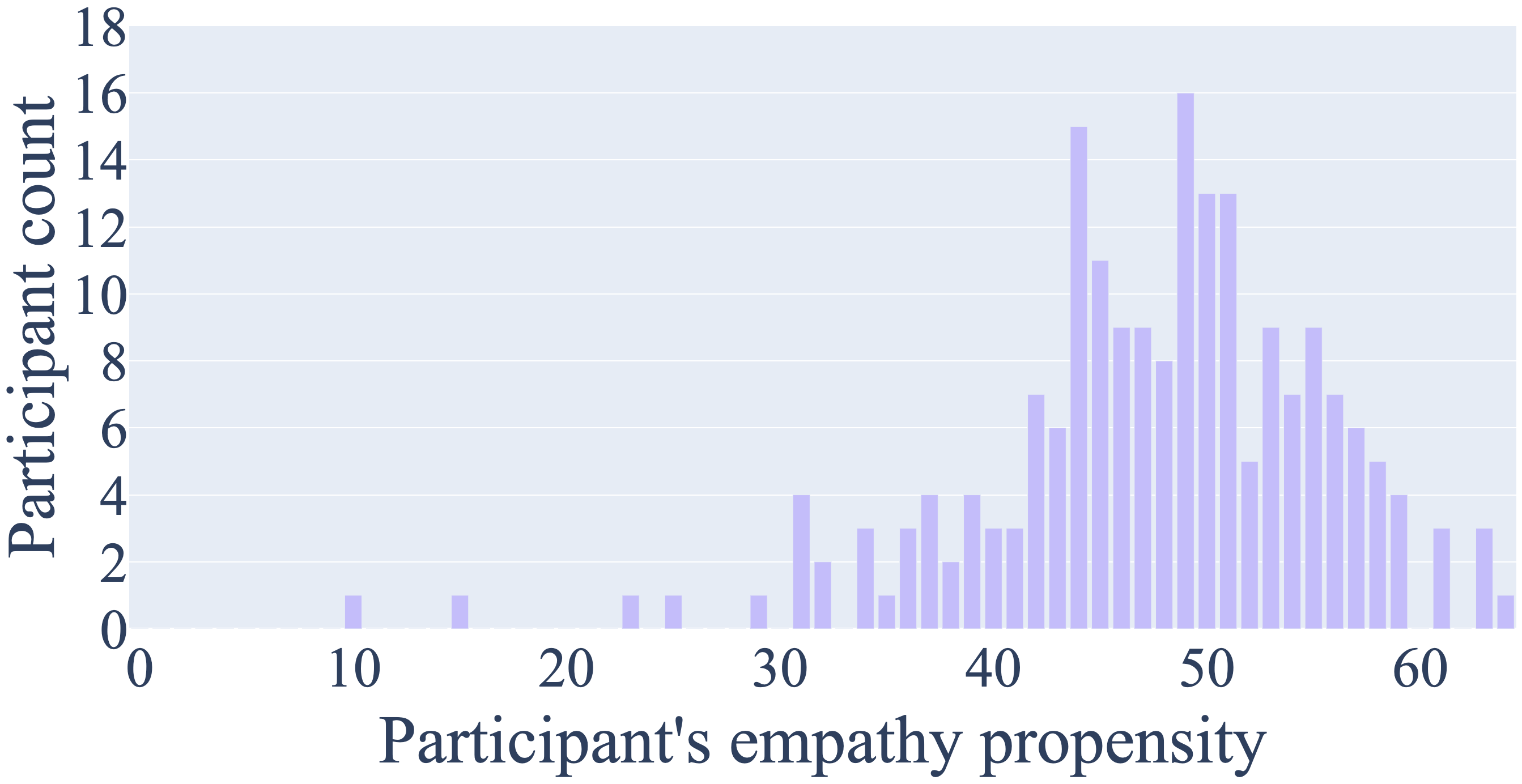}}}   

    \subfloat[Gemini-Pro responses raters]{{\includegraphics[width=0.33\textwidth]{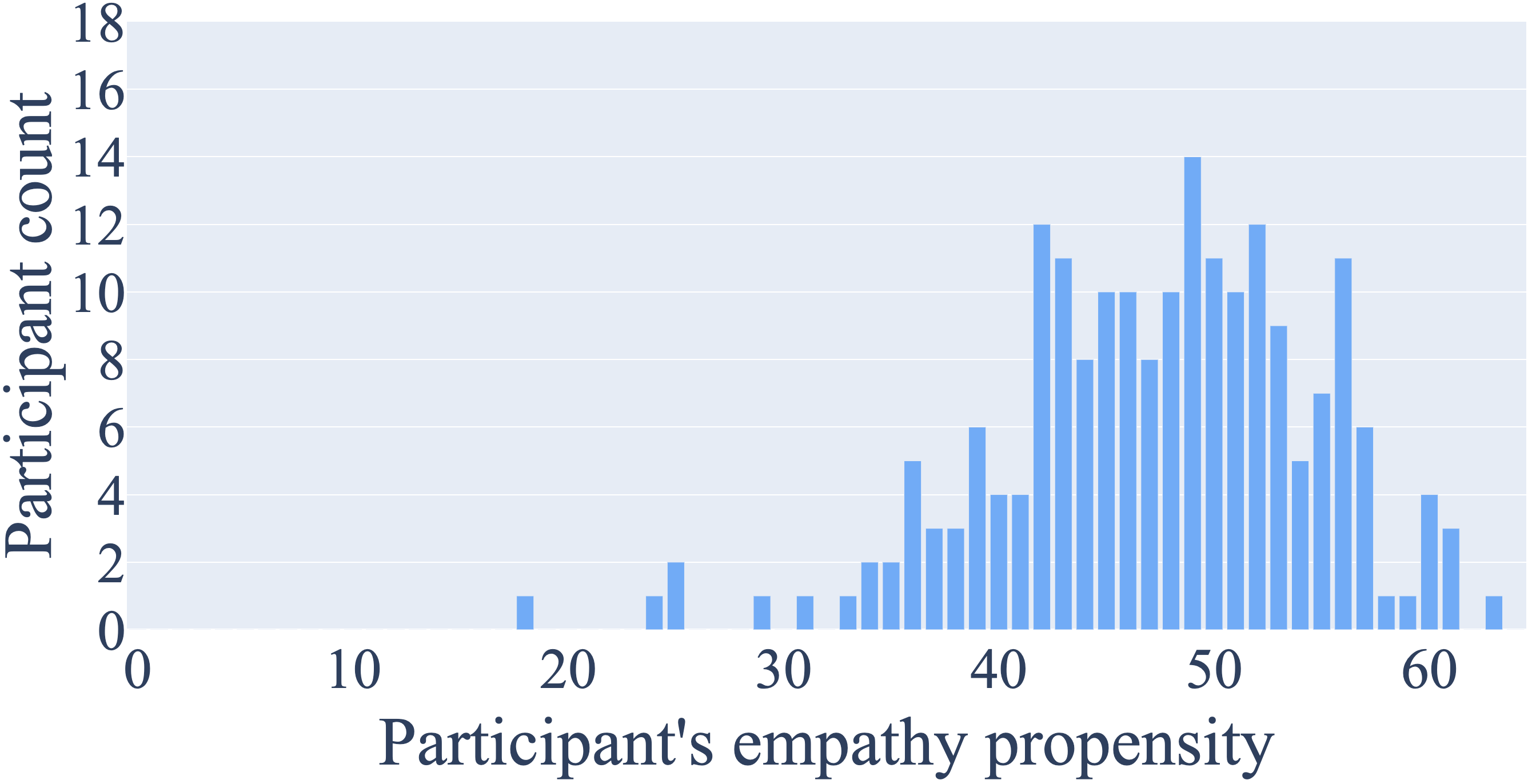}}}     
    \subfloat[Mixtral-8x7B responses raters]{{\includegraphics[width=0.33\textwidth]{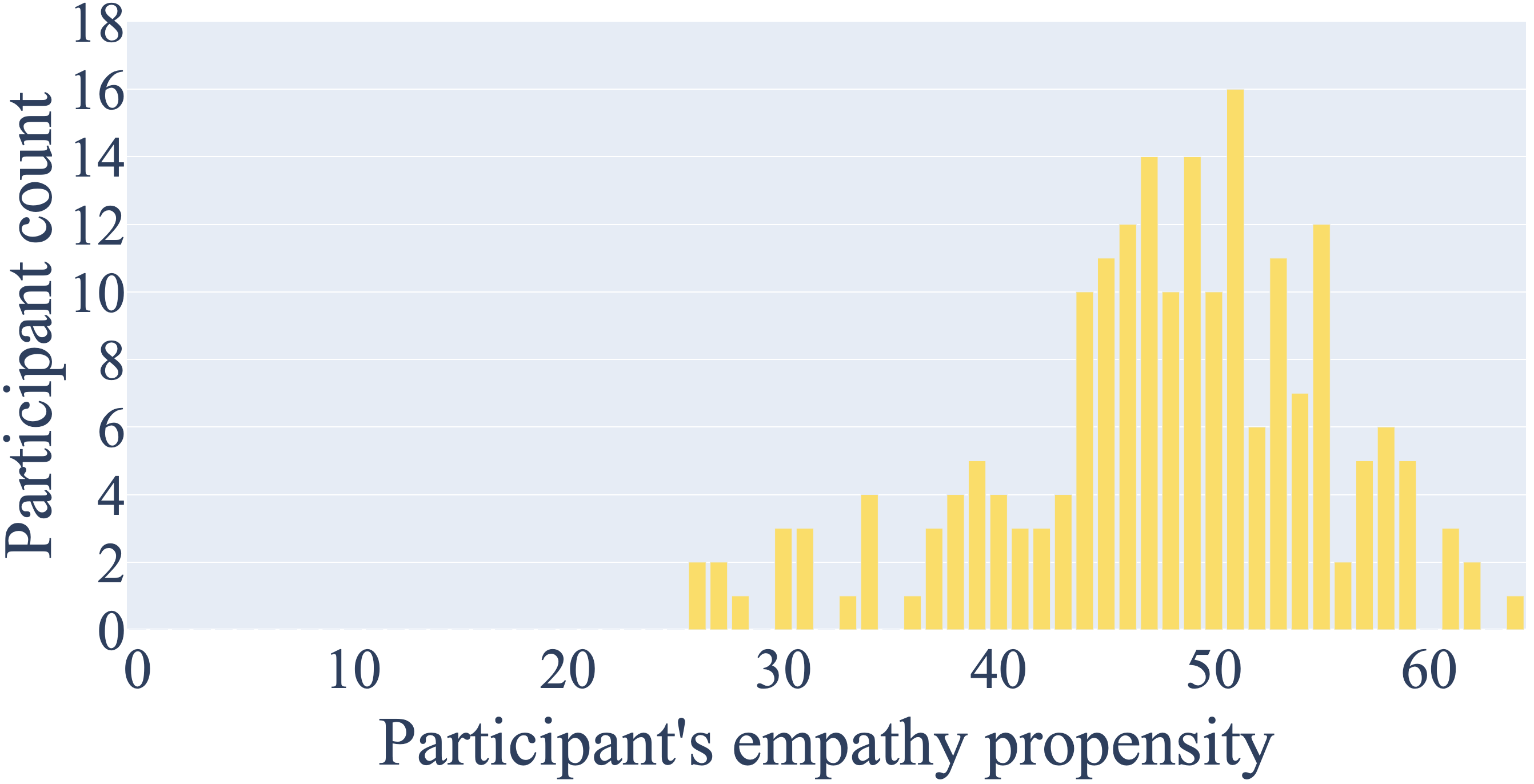}}}   
     
     \caption{The distributions of the participants' propensities to empathize across the five groups.}
     \label{fig:propensity}
\end{figure*}

Figure \ref{fig:propensity} shows the distributions of the participants' propensities to empathize across the five groups. It could be observed that they are more or less equally distributed across the three groups avoiding any biases in the results that might be caused by any inequal distribution of empathy propensities across the five groups. 

\section{Quality Analysis}
\label{app:quality}

\begin{figure}[H]
  \includegraphics[width=\linewidth]{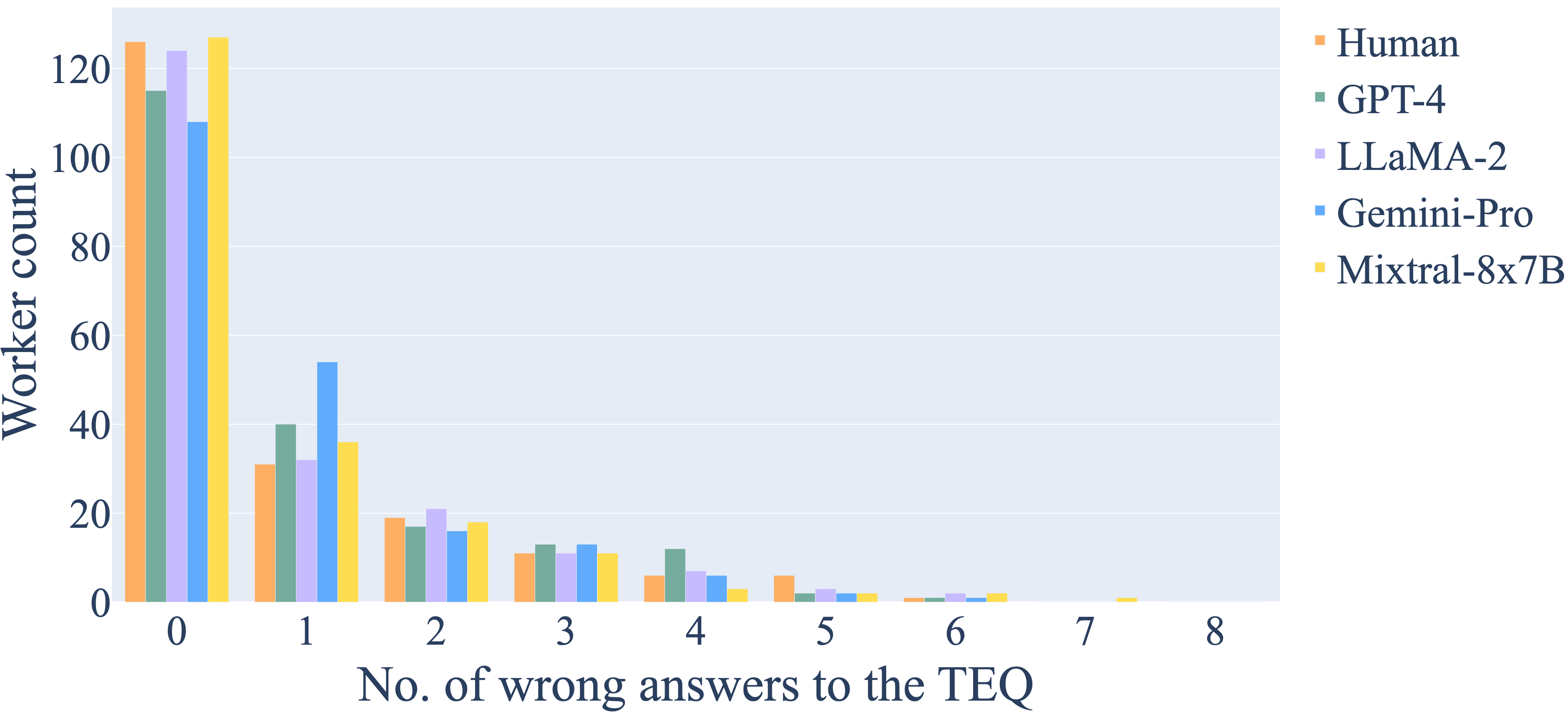}
  \caption{The number of reverse scale questions in the TEQ that were marked wrong by the participants rating the three response groups.}
  \label{fig:teq_histo}
\end{figure}

Figure \ref{fig:teq_histo} shows the number of reverse scale questions in the TEQ that were marked incorrect by the participants rating the three response groups. It was observed that 60\% of all participants did not get any reverse scale questions wrong and only 2.3\% of all participants got more than half of the reverse scale questions wrong. These statistics validate the quality of the workers recruited for the study. 

\begin{figure}[H]
  \includegraphics[width=\linewidth]{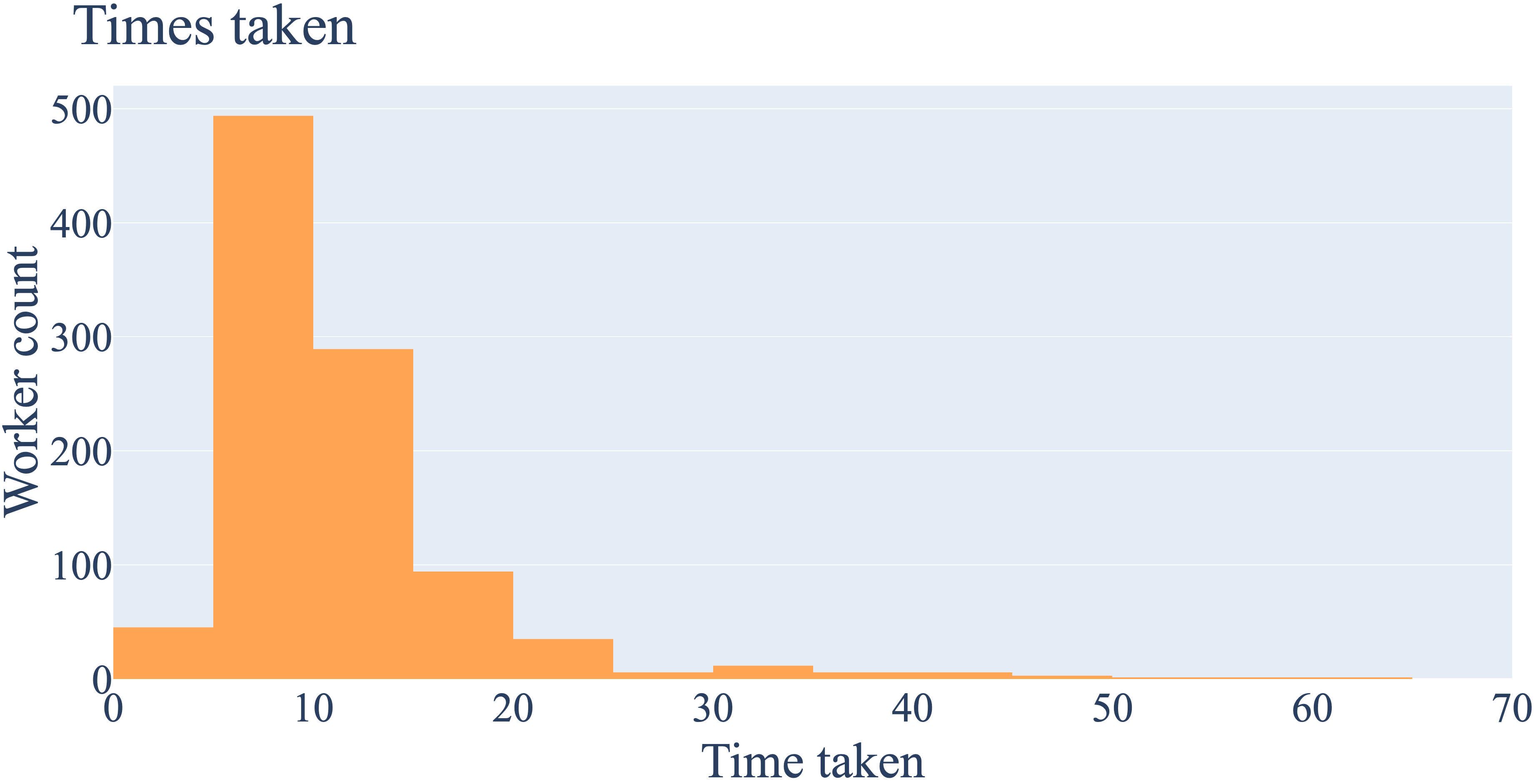}
  \caption{The histogram of times taken to complete the task by all participants.}
  \label{fig:time_histo}
\end{figure}

Further, Figure \ref{fig:time_histo} shows the histogram of times (in minutes) taken to complete the study. On average it took 11 minutes and 23 seconds to complete rating 10 responses, which was close to the average completion time of 15 minutes that we estimated before conducting the study. Only 4.53\% of all participants were observed to take less than 5 minutes to complete the study, which indicates that most of the participants took time to carefully read the instructions and respond to the questions attentively. 



\end{document}